\documentclass{article}

\PassOptionsToPackage{numbers,compress}{natbib}
\usepackage[nonatbib, final]{neurips_2022}
\usepackage{listings}

\usepackage[utf8]{inputenc} %
\usepackage{url}            %
\usepackage{booktabs}       %
\usepackage{amsfonts}       %
\usepackage{nicefrac}       %
\usepackage[table,xcdraw]{xcolor}
\usepackage{listings}
\usepackage{microtype}      %
\usepackage[table]{xcolor}
\usepackage{algorithm}
\usepackage[noend]{algpseudocode}

\usepackage{amsmath,amsfonts,bm}

\def\eqref#1{equation~\ref{#1}}

\def\1{\bm{1}}

\DeclareMathAlphabet{\mathsfit}{\encodingdefault}{\sfdefault}{m}{sl}
\SetMathAlphabet{\mathsfit}{bold}{\encodingdefault}{\sfdefault}{bx}{n}

\usepackage{graphicx}
\usepackage[font=small]{caption}
\usepackage{wrapfig}
\usepackage{booktabs}
\usepackage{multirow}
\usepackage{enumitem}
\usepackage{subcaption}
\usepackage{longtable}
\definecolor{citecolor}{HTML}{0071bc}
\usepackage{hyperref}
\hypersetup{breaklinks=true,colorlinks, citecolor=citecolor,linkcolor=black}
\usepackage[sort, square, compress]{natbib}
\setcitestyle{numbers}
\setcitestyle{citesep={,}}
\usepackage{multirow}
\usepackage{nicefrac}

\usepackage[OT1]{fontenc}
\usepackage{titlesec}
\usepackage{adjustbox}
\usepackage{amssymb}%
\usepackage{pifont}
\titleformat{\subsubsection}[runin]
  {\normalfont\normalsize\bfseries}{}{0.0em}{}

\titlespacing{\subsubsection}{0pt}{\parskip}{5pt}

\newcommand{\xxnote}[3]{}
\renewcommand{\xxnote}[3]{\color{#2}{#1: #3}}

\newcommand{\method}{Dobb·E}
\newcommand{\model}{HPR}
\newcommand{\stick}{Stick}
\newcommand{\dataset}{Homes of New York}
\newcommand{\datasetshort}{HoNY}
\newcommand{\numhomes}{10}
\newcommand{\numhomeexperiments}{109}
\newcommand{\numhomesuccess}{102}
\newcommand{\homesuccessrate}{81\%}
\newcommand{\datasettraj}{5620}
\newcommand{\datasethours}{13}
\newcommand{\datasethomes}{22}
\newcommand{\website}{https://dobb-e.com}
\title{On Bringing Robots Home}

\author{
    Nur Muhammad (Mahi) Shafiullah\thanks{Authors contributed equally.}\hspace{2pt}
    \thanks{Corresponding author, email: \texttt{mahi@cs.nyu.edu}}  \\ NYU
    \And Anant Rai$^*$ \\ NYU
    \And Haritheja Etukuru \\ NYU
    \And Yiqian Liu \\ NYU
    \AND Ishan Misra \\ Meta
    \And Soumith Chintala \\ Meta
    \And Lerrel Pinto \\ NYU
    \AND \textnormal{\url{\website}}
}

\begin{document}

\maketitle
\vspace{-2em}

\begin{abstract}
\label{abstract}
Throughout history, we have successfully integrated various machines into our homes.
Dishwashers, laundry machines, stand mixers, and robot vacuums are just a few recent examples.
However, these machines excel at performing only a single task effectively. 
The concept of a ``generalist machine'' in homes -- a domestic assistant that can adapt and learn from our needs, all while remaining cost-effective -- has long been a goal in robotics that has been steadily pursued for decades. 
In this work, we initiate a large-scale effort towards this goal by introducing \method{}, an affordable yet versatile general-purpose system for learning robotic manipulation within household settings.
\method{} can learn a new task with only five minutes of a user showing it how to do it, thanks to a demonstration collection tool (``The \stick'') we built out of cheap parts and iPhones. 
We use the \stick{} to collect \datasethours{} hours of data in \datasethomes{} homes of New York City, and train Home Pretrained Representations (\model{}).
Then, in a novel home environment, with five minutes of demonstrations and fifteen minutes of adapting the \model{} model, we show that \method{} can reliably solve the task on the Stretch, a mobile robot readily available on the market.
Across roughly 30 days of experimentation in homes of New York City and surrounding areas, we test our system in \numhomes{} homes, with a total of \numhomeexperiments{} tasks in different environments, and finally achieve a success rate of \homesuccessrate{}.
Beyond success percentages, our experiments reveal a plethora of unique challenges absent or ignored in lab robotics.
These range from effects of strong shadows to variable demonstration quality by non-expert users.
With the hope of accelerating research on home robots, and eventually seeing robot butlers in every home, we open-source \method{} software stack and models, our data, and our hardware designs.
\end{abstract}
\begin{figure}[h]
    \centering
    \includegraphics[width=0.95\linewidth]{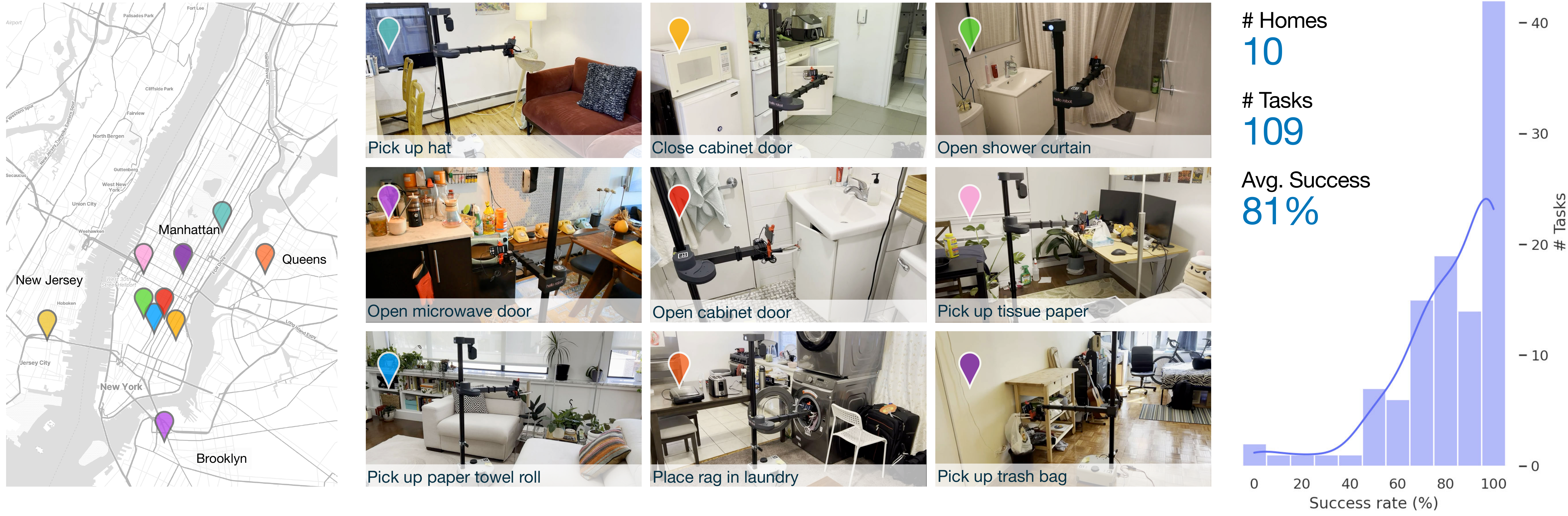}
    \label{fig:enter-label}
    \caption{ We present \method{}, a simple framework to train robots, which is then field tested in homes across New York City. In under 30 mins of training per task, \method{} achieves 81\% success rates on simple household tasks.}
\end{figure}
\vspace{-5em}

\newpage
\tableofcontents
\newpage

\section{Introduction}
\label{sec:intro}
Since our transition away from a nomadic lifestyle, homes have been a cornerstone of human existence. Technological advancements have made domestic life more comfortable, through innovations ranging from simple utilities like water heaters to advanced smart-home systems. However, a holistic, automated home assistant remains elusive, even with significant representations in popular culture~\cite{carper2019robots}.

Our goal is to build robots that perform a wide-range of simple domestic tasks across diverse real-world households. Such an effort requires a shift from the prevailing paradigm -- current research in robotics is predominantly either conducted in industrial environments or in academic labs, both containing curated objects, scenes, and even lighting conditions. In fact, even for the simple task of object picking~\cite{gupta2018robot} or point navigation~\cite{gervet2023navigating} performance of robotic algorithms in homes is far below the performance of their lab counterparts. If we seek to build robotic systems that can solve harder, general-purpose tasks, we will need to reevaluate many of the foundational assumptions in lab robotics.

\begin{figure}[b]
    \centering
    \includegraphics[width=\linewidth]{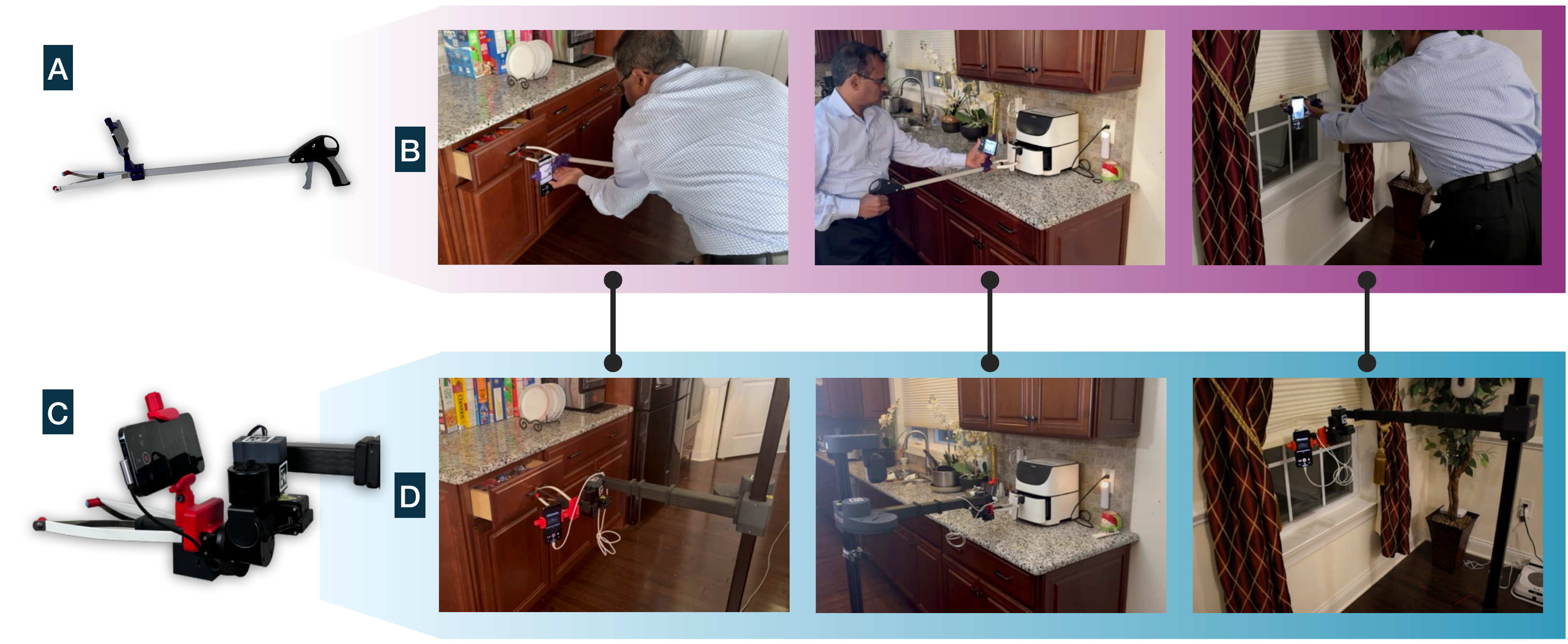}
    \caption{(A) We design a new imitation learning framework, starting with a data collection tool. (B) Using this data collection tool, users can easily collect demonstrations for household tasks. (C) Using a similar setup on a robot, (D) we can transfer those demos using behavior cloning techniques to real homes.}
    \label{fig:intro-figure}
\end{figure}

In this work we present \method{}, a framework for teaching robots in homes by embodying three core principles: efficiency, safety, and user comfort. For efficiency, we embrace large-scale data coupled with modern machine learning tools. For safety, when presented with a new task, instead of trial-and-error learning, our robot learns from a handful of human demonstrations. For user comfort, we have developed an ergonomic demonstration collection tool, enabling us to gather task-specific demonstrations in unfamiliar homes without direct robot operation. 

\begin{figure}[p!]
    \vspace{-4em}
    \centering
    \includegraphics[width=\linewidth,height=0.8\paperheight,keepaspectratio]{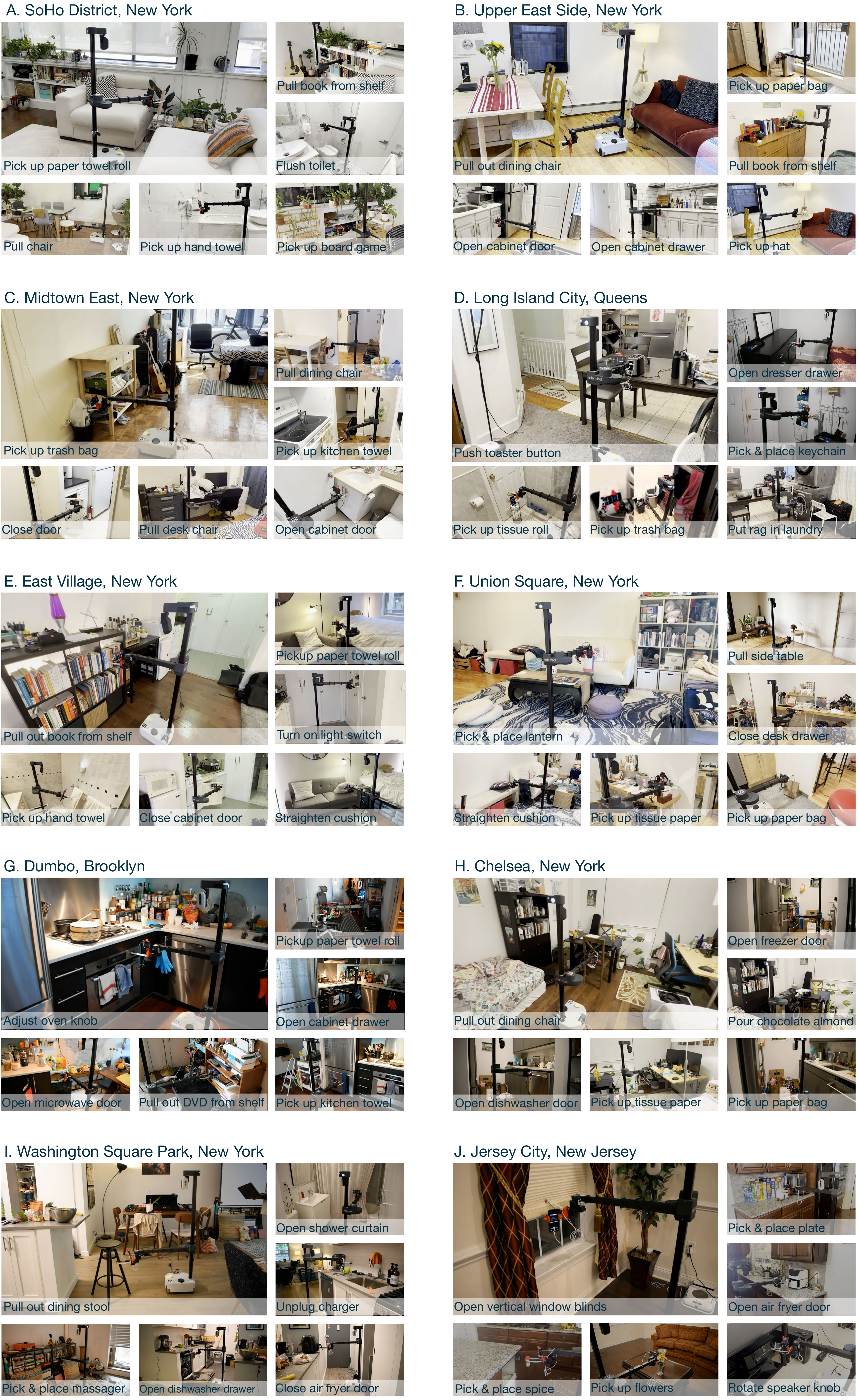}
    \caption{We ran experiments in a total of 10 homes near the New York City area, and successfully completed \numhomesuccess{} out of \numhomeexperiments{} tasks that we tried. The figure shows a subset of 60 tasks, 6 tasks from 10 homes each, from our home robot experiments using~\method{}.}
    \vspace{-3em}
    \label{fig:all-robots}
\end{figure}

Concretely, the key components of \method{} include:

\begin{itemize}[leftmargin=12pt]
    \item \textbf{Hardware}: The primary interface is our demonstration collection tool, termed the ``Stick.'' It combines an affordable reacher-grabber with 3D printed components and an iPhone. Additionally, an iPhone mount on the robot facilitates direct data transfer from the Stick without needing domain adaptation.

    \item \textbf{Pretraining Dataset}: Leveraging the Stick, we amass a \datasethours{} hour dataset called~\dataset{} (\datasetshort{}), comprising \datasettraj{} demonstrations from 216 environments in \datasethomes{} New York homes, bolstering our system's adaptability. This dataset serves to pretrain representation models for \method{}.

    \item \textbf{Models and algorithms}: Given the pretraining dataset we train a streamlined vision model, called Home Pretrained Representations (\model{}), employing cutting-edge self-supervised learning (SSL) techniques. For novel tasks, a mere 24 demonstrations sufficed to finetune this vision model, incorporating both visual and depth information to account for 3D reasoning.

    \item \textbf{Integration}: Our holistic system, encapsulating hardware, models, and algorithms, is centered around a commercially available mobile robot: Hello Robot Stretch~\cite{kemp2022stretch}.
\end{itemize}

We run \method{} across 10 homes spanning 30 days of experimentation, over which it tried \numhomeexperiments{} tasks and successfully learned \numhomesuccess{} tasks with performance $\geq 50\%$ and an overall success rate of ~\homesuccessrate{}. Concurrently, extensive experiments run in our lab reveals the importance of many key design decisions. Our key experimental findings are:

\begin{itemize}[leftmargin=12pt]
    \item \textbf{Surprising effectiveness of simple methods:} \method{} follows a simple behavior cloning recipe for visual imitation learning using a ResNet model~\cite{he2016deep} for visual representation extraction and a two-layer neural network~\cite{rosenblatt1958perceptron} for action prediction 
    (see Section~\ref{sec:technical-comp-and-method}). On average, only using 91 seconds of data on each task collected over five minutes, \method{} can achieve a \homesuccessrate{} success rate in homes (see Section~\ref{sec:experiments}).

    \item \textbf{Impact of effective SSL pretraining:} Our foundational vision model, \model{} trained on home data improves tasks success rate by at least $23\%$ compared to other foundational vision models~\cite{xiao2022mvp, nair2022r3m, majumdar2023vc1}, which were trained on much larger internet datasets (see Section~\ref{sec:ablations:representation}). 

    \item \textbf{Odometry, depth, and expertise:} The success of \method{} is heavily reliant on the \stick{} providing highly accurate odometry and actions from the iPhones' pose and position sensing, and depth information from the iPhone's Lidar. Ease of collecting demonsrations also makes iterating on research problems with the \stick{} much faster and easier (see Section~\ref{sec:ablations}).

    \item \textbf{Remaining challenges:} Hardware constraints such as the robot's force, reach, and battery life, limit tasks our robot can physically solve (see Section~\ref{sec:failures:hardware-limits}), while our policy framework suffers with ambiguous sensing and more complex, temporally-extended tasks (see Sections~\ref{sec:failures:temporal-dependencies},~\ref{sec:long-horizon}).
\end{itemize}

To encourage and support future work in home robotics, we have open-sourced our code, data, models, hardware designs, and are committed to supporting reproduction of our results. More information along with robot videos are available on our project website: \url{\website}.

\section{Technical Components and Method}
\label{sec:technical-comp-and-method}

\begin{figure}[t]
    \centering
    \includegraphics[width=\linewidth]{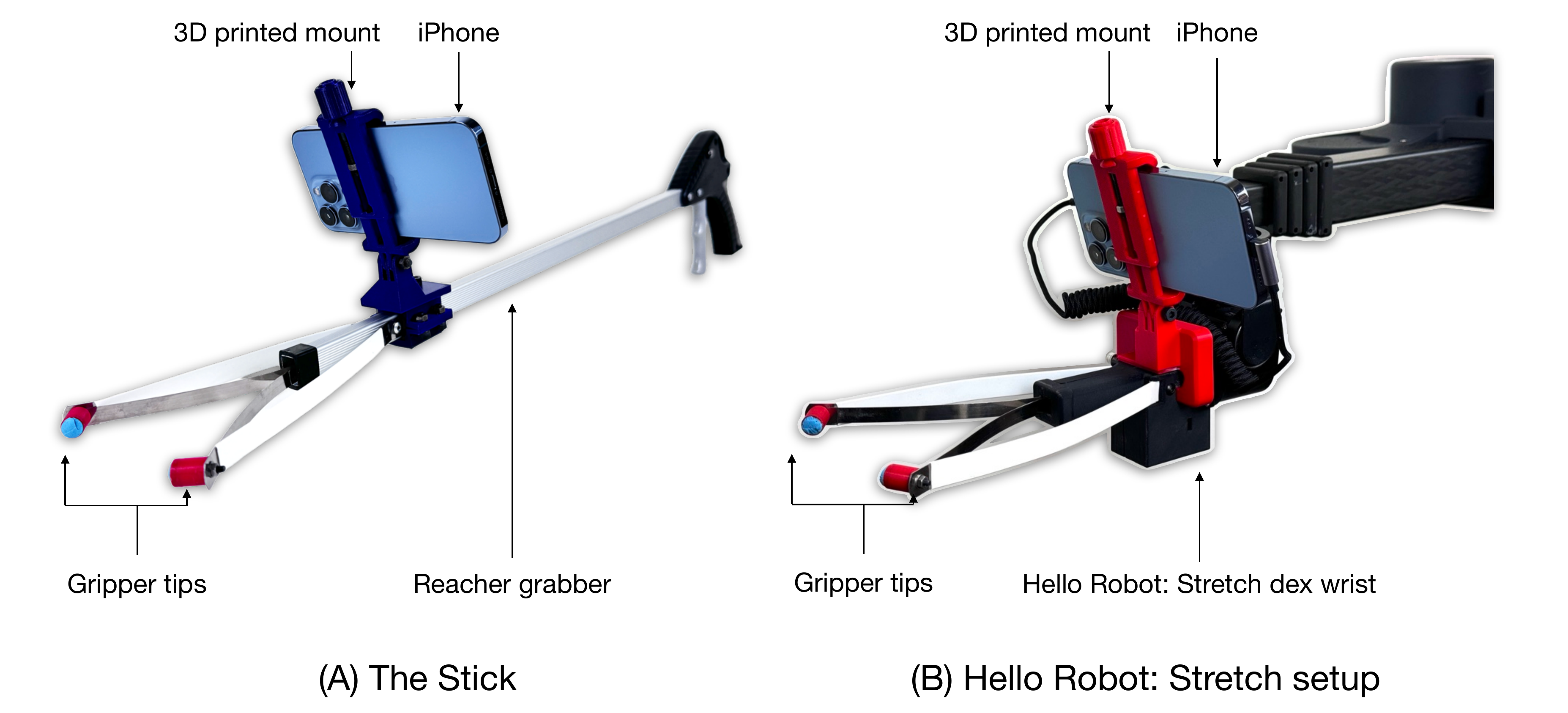}
    \caption{Photographs of our designed hardware, including the (A) \stick{} and the (B) identical iPhone mount for Hello Robot: Stretch wrist. From the iPhone's point of view, the grippers look identical between the two setups.}
    \label{fig:hardware}
\end{figure}

To create \method{} we partly build new robotic systems from first principles and partly integrate state-of-the-art techniques. In this section we will describe the key technical components in \method{}. To aid in reproduction of \method{}, we have open sourced all of the necessary ingredients in our work; please see Section~\ref{sec:reproducibility} for more detail.

At a high level, \method{} is an behavior cloning framework~\cite{atkeson1997robot}. Behavior cloning is a subclass of imitation learning, which is a machine learning approach where a model learns to perform a task by observing and imitating the actions and behaviors of humans or other expert agents. 
Behavior cloning involves training a model to mimic a demonstrated behavior or action, often through the use of labeled training data mapping observations to desired actions. In our approach, we pretrain a lightweight foundational vision model on a dataset of household demonstrations, and then in a new home, given a new task, we collect a handful of demonstrations and fine-tune our model to solve that task. However, there are many aspects of behavior cloning that we created from scratch or re-engineered from existing solutions to conform to our requirements of efficiency, safety, and user comfort.

Our method can be divided into four broad stages: (a) designing a hardware setup that helps us in the collection of demonstrations and their seamless transfer to the robot embodiment, (b) collecting data using our hardware setup in diverse households, (c) pretraining foundational models on this data, and (d) deploying our trained models into homes.

\subsection{Hardware Design}
\label{sec:hardware-design}
The first step in scaling robotic imitation to arbitrary households requires us to take a closer look at the standard imitation learning process and its inefficiencies. Two of the primary inefficiencies in current real-world imitation learning lay in the process of collecting the robotic demonstrations and transferring them across environments. 

\subsubsection{Collecting robot demonstrations}
\label{sec:collecting-robot-demo}
The standard approach to collect robot demonstrations in a robotic setup is to instrument the robot to pair it with some sort of remote controller device~\citep{mandlekar2018roboturk, arunachalam2023holodex}, a full robotic exoskeleton~\citep{fang2023airexo, devito, zhao2023wearable, ishiguro2020bilateral}, or simpler data collection tools~\citep{song2020grasping, young2020visual, pari2021surprising}. Many recent works have used a video game controller or a phone~\cite{mandlekar2018roboturk}, RGB-D cameras~\cite{arunachalam2022dexterous}, or virtual reality device~\citep{arunachalam2023holodex, guzey2023dexterity, guzey2023see} to control the robot. Other works~\citep{zhao2023aloha} have used two paired robots in a scene where one of the robots is physically moved by the demonstrator while the other robot is recorded by the cameras. However, such approaches are hard to scale up to households efficiently. Physically moving a robot is generally unwieldy, and for a home robotic task would require having multiple robots present at the site. Similarly, full exoskeleton based setups as shown in~\citep{fang2023airexo, zhao2023wearable, ishiguro2020bilateral} are also unwieldy in a household setting. Generally, the hardware controller approach suffers from inefficiency because the human demonstrators have to map the controller input to the robot motion. Using phones or virtual reality devices are more efficient, since they can map the demonstrators’ movements directly to the robot. However, augmenting these controllers with force feedback is nearly impossible, often leading users to inadvertently apply extra force or torque on the robot. Such demonstrations frequently end up being unsafe, and the generally accepted solution to this problem is to limit the force and torque users can apply; however, this often causes the robot to diverge from the human behavior.

In this project, we take a different approach by trying to combine the versatility of mobile controllers with the intuitiveness of physically moving the robot. Instead of having the users move the entire robot, we created a facsimile of the Hello Robot Stretch end-effector using a cheap~\$25 reacher-grabber stick that can be readily bought online, and augmented it ourselves with a 3D printed iPhone mount. We call this tool the “\stick{},” which is a natural evolution of tools used in prior work~\citep{young2021playful, pari2021surprising} (see Figure~\ref{fig:hardware}).

The \stick{} helps the user intuitively adapt to the limitations of the robot, for example by making it difficult to apply large amounts of force. Moreover, the iPhone Pro (version 12 or newer), with its camera setup and internal gyroscope, allows the \stick{} to collect RGB image and depth data at 30 frames per second, and its 6D position (translation and rotation). In the rest of the paper, for brevity, we will refer to the iPhone Pro (12 or later) simply as iPhone.

\subsubsection{Captured Data Modalities}
\label{sec:captured-data-modalities}
Our \stick{} collects the demonstration data via the mounted iPhone using an off-the-shelf app called Record3D. The Record3D app is able to save the RGB data at 1280$\times$720 pixels recorded from the camera, the depth data at 256$\times$192 pixels from the lidar sensor, and the 6D relative translation and rotation data from the iPhone’s internal odometry and gyroscope. We record this data at 30 FPS onto the phone and later export and process it.

\subsubsection{Robot Platform}
\label{sec:robot-platform}
All of our systems are deployed on the Hello Robot Stretch, which is a single-arm mobile manipulator robot already available for purchase on the open market. We use the Stretch RE1 version in all of our experiments, with the dexterous wrist attachment that confers 6D movement abilities on the robot. We chose this robot because it is cheap, lightweight--weighing just 51 pounds (23 kilograms)--and can run on a battery for up to two hours. Additionally, Stretch RE1 has an Intel NUC computer on-board which can run a learned policy at 30 Hz.

\subsubsection{Camera Mounts}
\label{sec:camera-mounts}
We create and use matching mounts on the \stick{} and the Hello Robot arm to mount our iPhone, which serves as the camera and the sensor in both cases. One of the main advantages of collecting our data using this setup is that, from the camera's point of view, the Stick gripper and the robot gripper looks identical, and thus the collected data and any trained representations and policies on such data can be directly transferred from the Stick to the robot. Moreover, since our setup operates with only one robot mounted camera, we don’t have to worry about having and calibrating a third-person, environment mounted camera, which makes our setup robust to general camera calibration issues and mounting-related environmental changes.

\subsubsection{Gripper Tips}
\label{sec:gripper-tips}
As a minor modification to the standard reacher-grabber as well as the Hello Robot Stretch end-effector, we replace the padded, suction-cup style tips of the grippers with small, cylindrical tips. This replacement helps our system manipulate finer objects, such as door and drawer handles, without getting stuck or blocked. In some preliminary experiments, we find that our cylindrical tips are better at such manipulations, albeit making pick-and-place like tasks slightly harder.

\subsection{Pretraining Dataset -- \dataset{}}
\label{sec:dataset}
\begin{figure}[t]
    \centering
    \includegraphics[width=\linewidth]{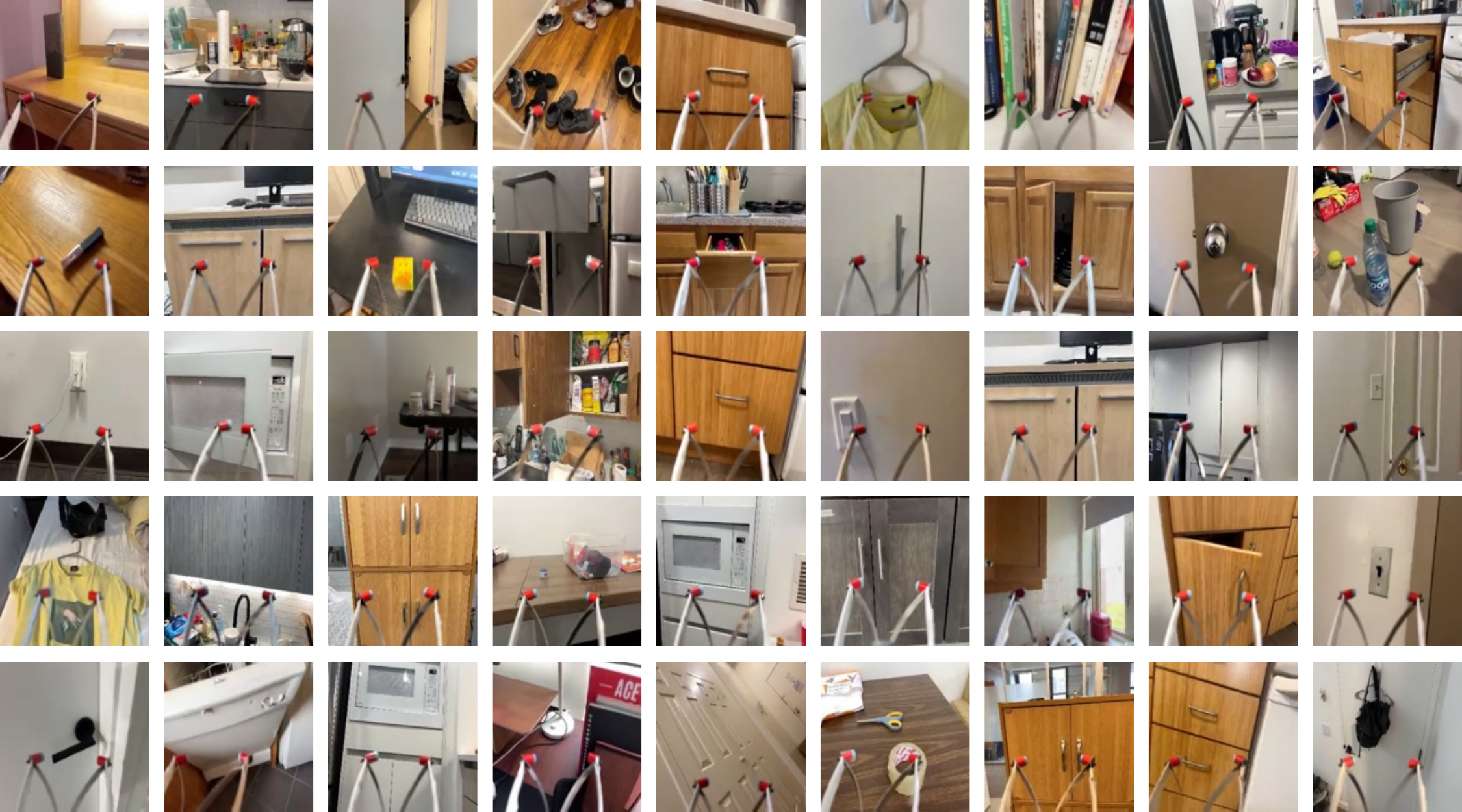}
    \caption{Subsample of 45 frames from \dataset{} dataset, collected using our \stick{} in \datasethomes{} homes.}
    \label{fig:dataset}
\end{figure}

With our hardware setup, collecting demonstrations for various household tasks becomes as simple as bringing the \stick{} home, attaching an iPhone to it, and doing whatever the demonstrator wants to do while recording with the Record3D app. To understand the effectiveness of the \stick{} as a data collection tool and give us a launching pad for our large-scale learning approach, we, with the help of some volunteers, collected a household tasks dataset that we call~\dataset{} (\datasetshort{}).

The~\datasetshort{} dataset is collected with the help of volunteers across \datasethomes{} different homes, and it contains \datasettraj{} demonstrations in \datasethours{} hours of total recording time and totalling almost 1.5 million frames. We asked the volunteers to focus on eight total defined broad classes of tasks: switching button, door opening, door closing, drawer opening, drawer closing, pick and place, handle grasping, and play data. For the play data, we asked the volunteers to collect data from doing anything arbitrary around their household that they would like to do using the stick. Such playful behavior has in the past proven promising for representation learning purposes~\cite{young2021playful, guzey2023dexterity}.

We instructed our volunteers to spend roughly 10 minutes to collect demonstrations in each ``environment'' or scene in their household. However, we did not impose any limits on how many different tasks they can collect in each home, nor how different each ``environment'' needs to be across tasks. Our initial demonstration tasks were chosen to be diverse and moderately challenging while still being possible for the robot.

\begin{figure}[t]
    \centering
    \includegraphics[width=\linewidth]{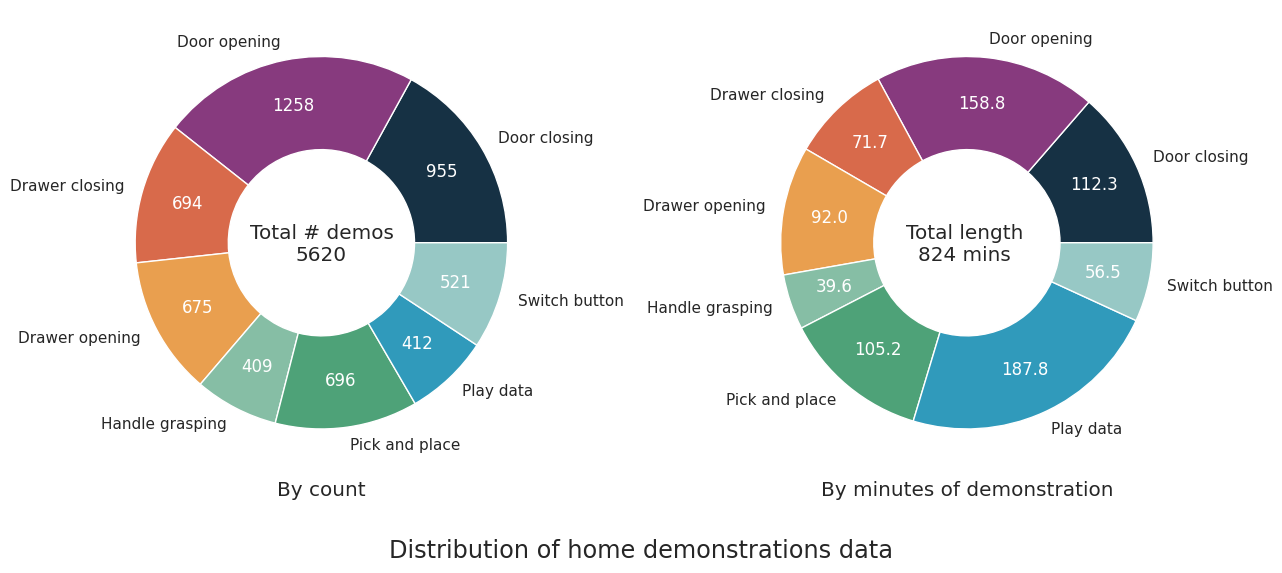}
    \caption{Breakdown of~\dataset{} dataset by task: on the left, the statistics is shown by number of demonstrations, and on the right, the breakdown is shown by minutes of demonstration data collected.}
    \label{fig:dataset-breakdown-task}
\end{figure}

In Figure~\ref{fig:dataset-breakdown-task}, we can see a breakdown of the dataset by the number of frames belonging to each broad class of tasks. As we can see, while there is some imbalance between the number of frames in each task, they are approximately balanced.

\begin{figure}[t]
    \centering
    \includegraphics[width=\linewidth]{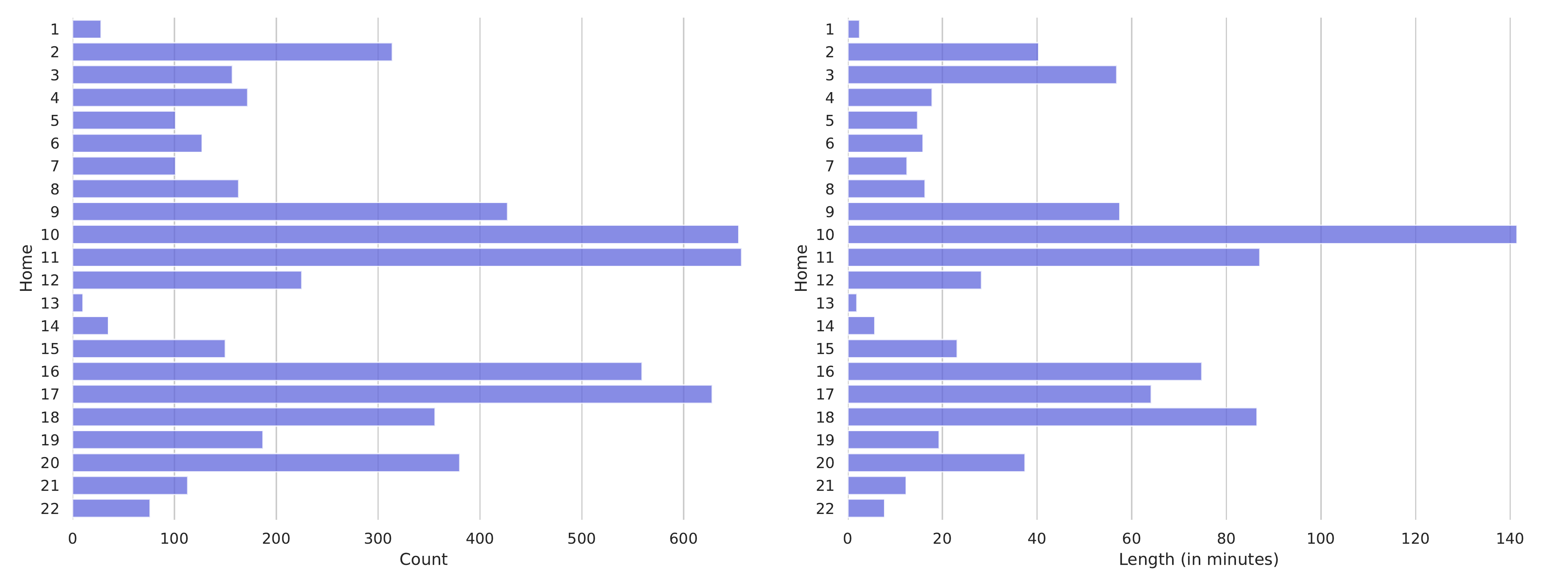}
    \caption{Breakdown of our collected dataset by homes. On the left, the statistics are shown by number of demonstrations, and on the right, the breakdown is shown by minutes of demonstration data collected. The Y-axis is marked with the home ID.}
    \label{fig:dataset-breakdown-homes}
\end{figure}
Moreover, our dataset contains a mixture of a diverse number of homes, as shown in Figure~\ref{fig:dataset-breakdown-homes}, with each home containing 67K frames and 255 trajectories on average.

\subsubsection{Gripper Data}
\label{sec:gripper-data}
While the iPhone can give us the pose of the end-effector, there is no way to trivially get the open or closed status of the gripper itself. To address this, we trained a model to track the gripper tips. We extracted 500 random frames from the dataset and marked the two gripper tip positions in pixel coordinates on those frames. We trained a gripper model on that dataset, which is a 3-layer ConvNet that tries to predict the distance between the gripper tips as a normalized number between 0 and 1. This model, which gets a 0.035 MSE validation error (on a scale from 0-1) on a heldout evaluation set, is then used to label the rest of the frames in the dataset with a gripper value between 0 and 1.

\subsubsection{Dataset Format}
\label{sec:dataset-format}
 As mentioned in the previous section, we collect the RGB and depth data from the demonstration, as well as the 6D motion of the stick, at 30 Hz. For use in our models, we scale and reshape our images and depths into 256$\times$256 pixels. For the actions, we store the absolute 6D poses of the iPhone at 30 Hz. During model training or fine-tuning, we calculate the relative pose change as the action at the desired frequency during runtime.

\subsubsection{Dataset Quality Control}
\label{sec:quality-control}
 We manually reviewed the videos in the dataset to validate them and filter them for any bad demonstrations, noisy actions, and any identifying or personal information. We filtered out any videos that were recorded in the wrong orientation, as well as any videos that had anyone's face or fingers appearing in them.

\subsubsection{Related Work}
\label{sec:dataset-related-work}
Collecting large robotic manipulation datasets is new. Especially in recent years, there have been a few significant advances in collecting large datasets for robotics \citep{brohan2022rt, jang2021bc, fang2023rh20t, mandlekar2019scaling, ebert2021bridge, walke2023bridgedata, gupta2018robot, jiang2011efficient, Pinto2015SupersizingSL, bohg2015dataset, Mahler2017DexNet2D, depierre2018jacquard, levine2018learning, kalashnikov2018qt, Brahmbhatt2019, fang2020graspnet, acronym2020, bousmalis2018grasping, zhu2023fanuc, yu2016more,finn2017deep,ebert2018visual,dasari2019robonet, kalashnikov2021mt, ebert2021bridge, robomimic2021, brohan2023rt2, lynch2023interactive, roboagent, heo2023furniturebench}. While our dataset is not as large as the largest of them, it is is unique in a few different ways. Primarily, our dataset is focused on household interactions, containing 22 households, while most datasets previously were collected in laboratory settings. Secondly, we collect first-person robotic interactions, and are thus inherently more robust to camera calibration issues which affect previous datasets \citep{sharma2018multiple, mandlekar2018roboturk, cabi2019scaling, kalashnikov2021mt, jang2021bc, bharadhwaj2023roboagent}. Thirdly, using an iPhone gives us an advantage over previous work that used cheap handheld tools to collect data \citep{song2020grasping, young2020visual, pari2021surprising} since we can extract high quality action information quite effortlessly using the onboard gyroscope. Moreover, we collect and release high quality depth information from our iPhone, which is generally rare for standard robotic datasets. The primary reason behind collecting our own dataset instead of using any previous dataset is because we believe in-domain pretraining to be a key ingredient for generalizable representations, which we empirically verify in section~\ref{sec:ablations:representation} by comparing with previously released general-purpose robotic manipulation focused representation models. A line of work that may aid in future versions of this work are collections of first-person non-robot household videos, such as~\cite{damen2018epickitchen, grauman2022ego4d, somasundaram2023projectaria}, where they can complement our dataset by augmenting it with off-domain information.

\newcommand{\cmark}{\ding{51}}
\newcommand{\xmark}{\ding{55}}

\begin{table}[t]
    \centering
    \caption{While previous datasets focused on the number of manipulation trajectories, we instead focus on diverse scenes and environments. As a result, we end up with a dataset that is much richer in interaction diversity.}
    \vspace{0.2cm}
    \begin{adjustbox}{center}
        \footnotesize
        \begin{tabular}{l c c c c c >{\arraybackslash} m{2cm}}
            \toprule
            Dataset                               & \# Traj. & \# Env. & \# Homes  & Public Data & Public Robot & Collection                               \\
            \midrule
            MIME \cite{sharma2018multiple}        & 8.30k          & 1    & -   & \cmark      & \cmark       & human                                    \\
            RoboTurk \cite{mandlekar2018roboturk} & 2.10k            & 1       & - & \cmark      & \cmark       & human                                    \\
            Learning in Homes \cite{gupta2018robot}      & 28k            & 9      & 9 & \cmark      & \cmark       & scripted                                 \\
            MT-Opt \cite{kalashnikov2021mt}    & 800k             & 1       & - & \xmark      & \cmark       & \mbox{scripted \& learned}               \\
            BC-Z \cite{jang2021bc}                & 26.0k             & 1       & - & \cmark      & \xmark       & human                                    \\
            RT-1 \cite{brohan2022rt}              & 130k              & 3       & - & \cmark      & \xmark       & human                                    \\
            RH20T \cite{fang2023rh20t}            & 110k             & 50     & 10 & \cmark      & \cmark       & human                                    \\
            RoboSet \cite{bharadhwaj2023roboagent}              & 98.5k              & 11       & - & \cmark      & \cmark       & \mbox{scripted \& human}                                    \\
            BridgeData v2~\cite{walke2023bridgedata}               & 60.1k     & 24 & - & \cmark      & \cmark       & \mbox{human \& scripted} \\
            \midrule
            \datasetshort{} (Us)               & 5.6k    & \textbf{216} & \textbf{22} & \cmark      & \cmark       & human \\
            \bottomrule                                                                                                                                             \\
        \end{tabular}
    \end{adjustbox}
    \captionsetup{width=\textwidth}
    \label{tab:comparison}
\end{table}

\subsection{Policy Learning with Home Pretrained Representations}
\label{sec:policy-learning}

\begin{figure}[t]
    \vspace{-1em}
    \centering
    \includegraphics[width=\linewidth]{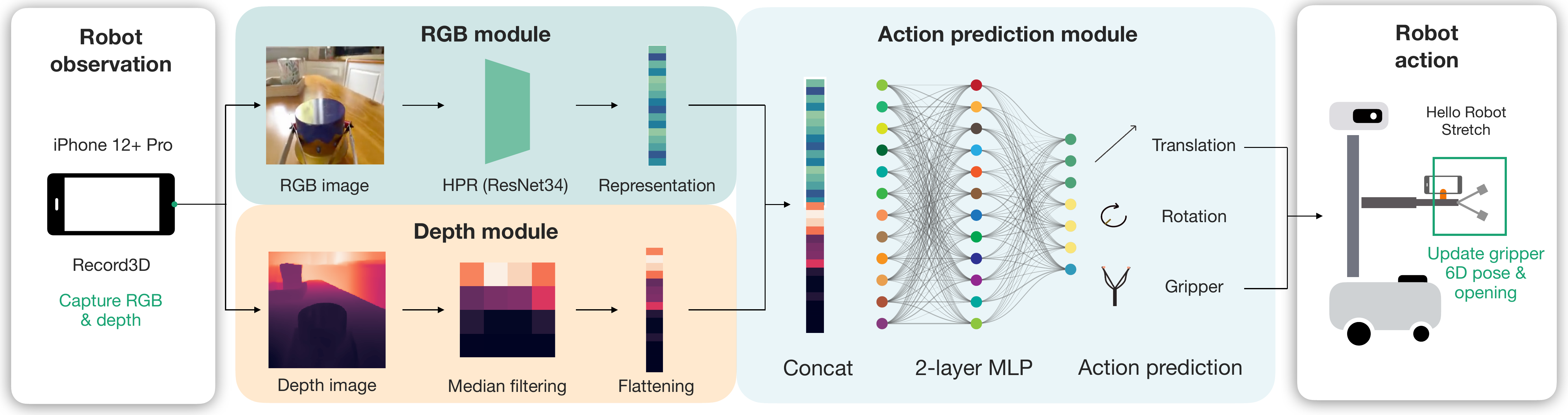}
    \caption{Fine-tuning the pretrained HPR model to learn a model that maps from the robot's RGB and depth observations into robot actions: 6D relative pose and the gripper opening.}
    \label{fig:methods}
    \vspace{-1em}
\end{figure}

With the diverse home dataset, our next step in the process is to train a foundational visual imitation model that we can easily modify and deploy in homes. To keep our search space small, in this work we only consider simple visual imitation learning algorithms that only consider a single step at a time. While this inevitably limits the capabilities of our system, we leave temporally extended policies as a future direction we want to explore on home robots. Our policy is built of two simple components: a visual encoder and a policy head.

\subsubsection{Visual Encoder Learning}
\label{sec:hpr-learning}
We use a ResNet34 architecture as a base for our primary visual encoder. While there are other novel architectures that were developed since ResNet34, it satisfies our need for being performant while also being small enough to run on the robot’s onboard computer. We pretrain our visual encoder on our collected dataset with the MoCo-v3 self-supervised learning algorithm for 60 epochs. We call this model the Home Pretrained Representation (HPR) model, based on which all of our deployed policies are trained. We compare the effects of using our own visual encoder vs. a pretrained visual encoder trained on different datasets and algorithms, such as R3M~\cite{nair2022r3m}, VC1~\cite{majumdar2023vc1}, and MVP~\cite{xiao2022mvp}, or even only pretraining on ImageNet-1K~\cite{deng2009imagenet}, in Section~\ref{sec:ablations:representation}.

\subsubsection{Downstream Policy Learning}
\label{sec:downstream-policy-learning}
On every new task, we learn a simple manipulation policy based on our visual encoder and the captured depth values. For the policy, the input space is an RGB-D image (4 channels) with shape 256$\times$256 pixels, and the output space is a 7-dimensional vector, where the first 3 dimensions are relative translations, next 3 dimensions are relative rotations (in axis angle representation), and the final dimension is a gripper value between 0 and 1. Our policy is learned to predict an action at 3.75 Hz, since that is the frequency with which we subsample our trajectories.

The policy architecture simply consists of our visual representation model applied to the RGB channels in parallel to a median-pooling applied on the depth channel, followed by two fully connected layers that project the 512 dimensional image representation and 512 dimensional depth values down to 7 dimensional actions. During this supervised training period where the network learns to map from observation to actions, we do not freeze any of the parameters, and train them for 50 epochs with a learning rate of $3\times 10^{-5}$. We train our network with a mean-squared error (MSE) loss, and normalize the actions per axis to have zero mean and unit standard deviation before calculating the loss.

Our pretrained visual encoders and code for training a new policy on your own data is available open-source with a permissive license. Please see Section~\ref{sec:reproducibility} for more details.

\subsubsection{Related Work}
\label{sec:finetuning-related-work}
While the pretraining-finetuning framework has been quite familiar in other areas of Machine Learning such as Natural Language~\citep{devlin2018bert, brown2020gpt3} and Computer Vision~\citep{he2016deep, oquab2023dinov2}, it has not caught on in robot learning as strongly. Generally, pretraining has taken the form of either learning a visual representation~\citep{brandfonbrener2023inverse, nair2022r3m, majumdar2023vc1, xiao2022mvp, pari2021surprising, young2021playful, radosavovic2022real, ma2022vip, karamcheti2023language, mu2023ec2, bahl2023affordances} or learning a Q-function~\cite{kumar2022pre, herzog2023deep} which is then used to figure out the best behavior policy. In this work, we follow the first approach, and pretrain a visual representation that we fine-tune during deployment. While there are recent large-scale robotic policy learning approaches~\cite{brohan2022rt, brohan2023rt2, padalkar2023open}, the evaluation setup for such policies generally have some overlap with the (pre-)training data. This work, in contrast, focuses on entirely new households which were never seen during pretraining.

\subsection{Deployment in Homes}
\label{sec:deployment}

\begin{figure}[t]
    \centering
    \includegraphics[width=\linewidth]{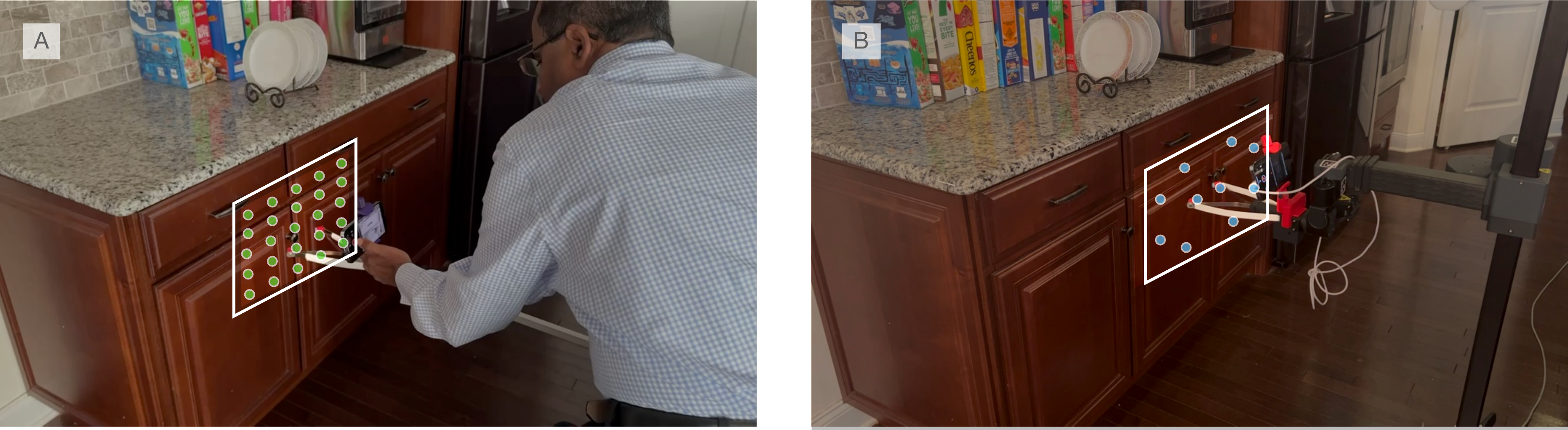}
    \caption{(a) The data collection grid: the demonstrator generally started data collection from a 5$\times$5 or 4$\times$6 grid of starting positions to ensure diversity of the collected demos. (b) To ensure our policies generalize to different starting positions, we start the robot policy roll-outs from 10 pre-scheduled starting positions.}
    \label{fig:data-collection-grid}
    \vspace{-1em}
\end{figure}

Once we have our \stick{} to collect data, the dataset preparation script, and the algorithm to fine-tune our pretrained model, the final step is to combine them and deploy them on a real robot in a home environment. In this work, we focus on solving tasks that mostly involve manipulating the environment, and thus we assume that the robot has already navigated to the task space and is starting while facing the task target (which for example could be an appliance to open or an object to manipulate).

\subsubsection{Protocol for Solving Home Tasks}
\label{sec:deployment-protocol}
In a novel home, to solve a novel task, we start by simply collecting a handful of demonstrations on the task. We generally collect 24 new demonstrations as a rule of thumb, which our experiments show is sufficient for simple, five second tasks. In practice, collecting these demos takes us about five minutes. However, some environments take longer to reset, in which case collecting demonstrations may also take longer. To confer some spatial generalization abilities to our robot policy, we generally collect the data starting from a variety of positions in front of the task setup, generally in a small 4$\times$6 or 5$\times$5 grid (Figure~\ref{fig:data-collection-grid}).

\subsubsection{Policy Training Details}
\label{sec:finetuning-during-deployment}
Once the data is collected, it takes about 5 minutes to process the data from R3D files into our dataset format. From there, for 50 epochs of training it takes about 20 minutes on average on a modern GPU (RTX A4000). As a result, on average, within 30 minutes from the start of the data collection, we end up with a policy that we can deploy on the robot.

\subsubsection{Robot Execution Details}
\label{sec:robot-execution}
We deploy the policy on the robot by running it on the robot’s onboard Intel NUC computer. We use the iPhone mounted on the arm and the Record3D app to stream RGB-D images via USB to the robot computer. We run our policy on the input images and depth to get the predicted action.
We use a PyKDL based inverse kinematics solver to execute the predicted relative action on the robot end-effector.
Since the model predicts the motion in the camera frame, we added a joint in the robot’s URDF for the attached camera, and so we can directly execute the predicted action without exactly calculating the transform from the camera frame to the robot end-effector frame.
For the gripper closing, we binarize the predicted gripper value by applying a threshold that can vary between tasks.
We run the policy synchronously on the robot by taking in an observation, commanding the robot to execute the policy-predicted action, and waiting until robot completes the action to take in the next observation.
For our evaluation experiments we generally use 10 initial starting positions for each robot task (Figure~\ref{fig:data-collection-grid}~(b)). These starting positions vary our robot gripper’s starting position in the vertical and horizontal directions. Between each of these 10 trials, we manually reset the robot and the environment.

\subsubsection{Related Work}
\label{sec:related-work}
While the primary focus of our work is deploying robots in homes, we are not the first one to do so.
The most popular case would be commercial robots such as Roomba~\cite{jones2006irobot} from iRobot or Astro~\cite{dempsey2023astro} from Amazon.
While impressive as a commercial product, such closed-source robots are not conducive to scientific inquiry and are difficult to build upon as a community.
Some application of robots in home includes early works such as~\cite{nguyen2014autonomously} exploring applications of predefined behaviors in homes,~\cite{bhattacharjee2016data, bhattacharjee2018multimodal} exploring tactile perception in homes, or ~\cite{gupta2018robot} exploring the divergence between home and lab data.
More recently, ObjectNav, i.e. navigating to objects in the real world~\cite{gervet2023navigating} has been studied by taking robots to six different houses.
While~\cite{gervet2023navigating} mostly experimented on short-term rental apartments and houses, we focused on homes that are currently lived in where cluttered scenes are much more common.
There have been other works such as~\cite{bahl2022whirl,shah2022viking} which focus on ``in the wild'' evaluation.
However, evaluation-wise, such works have been limited to labs and educational institutions~\cite{bahl2022whirl}, or have focused on literal ``in the wild'' setups such as cross-country navigation~\cite{shah2022viking}.

\section{Experiments}
\label{sec:experiments}
We experimentally validated our setup by evaluating it across \numhomes{} households in the New York and New Jersey area on a total of \numhomeexperiments{} tasks. On these \numhomeexperiments{} tasks, the robot gets an 81\% success rate, and can complete \numhomesuccess{} tasks with at least even odds. Alongside these household experiments, we also set up a “home” area in our lab, with a benchmark suite with 10 tasks that we use to run our baselines and ablations. Note that none of our experiments overlapped with the environments on which our HoNY dataset was collected to ensure that the experimental environments are novel.

\subsection{List of Tasks in Homes}
\label{sec:home-task-list}
In Table~\ref{tab:all-tasks} we provide an overview of the \numhomeexperiments{} tasks that we attempted in the~\numhomes{} homes, as well as the associated success rate on those tasks. Video of all \numhomeexperiments{} tasks can also be found on our website:~\url{\website /\#videos}.

\definecolor{darkgray}{cmyk}{0.68, 0.28, 0, 0.73}
\begin{longtable}{| l c c c c |} 

\caption{A list of all tasks in the home enviroments, along with their categories and success rates out of 10 trials.} \\
\hline
\rowcolor{darkgray}\color{white}
ID & \color{white}Home & \color{white}Task Description                         & \color{white}Success $\nicefrac{\cdot}{10}$ & \color{white}Task Category     \\ 
\hline
\endfirsthead
\hline
\rowcolor{darkgray}\color{white}
ID & \color{white}Home & \color{white}Task Description                         & \color{white}Success $\nicefrac{\cdot}{10}$ & \color{white}Task Category      \\ 
\hline
\endhead
\hline
\multicolumn{5}{|r|}{\small{Continued on the next page}} \\
\hline
\endfoot
\hline
\endlastfoot
1    & 1    & Door closing: Brown Cabinet             & 10      & Door closing          \\
2    & 1    & Drawer closing: Brown Drawer            & 10      & Drawer closing        \\
3    & 1    & Drawer Opening: Brown Drawer            & 10      & Drawer opening        \\
4    & 1    & Pick up: Plastic Plate                  & 9       & Misc object pickup    \\
5    & 1    & Pick up: Flowers                        & 3       & Misc object pickup    \\
6    & 1    & Pick and Place: Spices                  & 6       & 6D pick \& place        \\
7    & 1    & Pouring: translucent cup + marshmallows & 10      & Pouring               \\
8    & 1    & Air Fryer Opening                       & 10      & Air-fryer opening     \\
9    & 1    & Air Fryer Closing                       & 10      & Air-fryer closing     \\
10   & 1    & Knob Turning                            & 8       & Knob turning          \\
11   & 1    & Vertical Blinds Opening                 & 2       & Random                \\
12   & 1    & Horizontal Blinds Opening               & 10      & Random                \\
\rowcolor[HTML]{D9D9D9} 
13   & 2    & Sideways washing machine door           & 8       & Door opening          \\
\rowcolor[HTML]{D9D9D9} 
14   & 2    & Dresser drawer                          & 8       & Drawer opening        \\
\rowcolor[HTML]{D9D9D9} 
15   & 2    & Placing a rag in laundry                & 7       & 6D pick \& place        \\
\rowcolor[HTML]{D9D9D9} 
16   & 2    & Picking and placing a keyring           & 9       & 6D pick \& place        \\
\rowcolor[HTML]{D9D9D9} 
17   & 2    & Pouring: transparent cup                   & 5       & Pouring               \\
\rowcolor[HTML]{D9D9D9} 
18   & 2    & Trash pickup                            & 9       & Bag pickup            \\
\rowcolor[HTML]{D9D9D9} 
19   & 2    & Toilet paper unloading                  & 8       & Random                \\
\rowcolor[HTML]{D9D9D9} 
20   & 2    & Toaster button pressing                 & 1       & Random                \\
21   & 3    & Dishwasher drawer opening               & 8       & Drawer opening        \\
22   & 3    & Cat massager pick and place (onto book) & 7       & 6D pick \& place        \\
23   & 3    & Rattatoullie pick and place             & 5       & 6D pick \& place        \\
24   & 3    & Air fryer opening                       & 0       & Air-fryer opening     \\
25   & 3    & Air fryer closing                       & 10      & Air-fryer closing     \\
26   & 3    & Chair pulling                           & 10      & Chair pulling         \\
27   & 3    & Light switch new demos                  & 8       & Light switch          \\
28   & 3    & Unplugging                              & 10      & Unplugging            \\
29   & 3    & Towel pickup                            & 7       & Towel pickup          \\
30   & 3    & Kettle switch                           & 0       & Random                \\
31   & 3    & Shower curtains                         & 6       & Random                \\
\rowcolor[HTML]{D9D9D9} 
32   & 4    & Cabinet door closing                    & 10      & Door closing          \\
\rowcolor[HTML]{D9D9D9} 
33   & 4    & Closet door opening                     & 7       & Door opening          \\
\rowcolor[HTML]{D9D9D9} 
34   & 4    & Freezer door opening                    & 9       & Door opening          \\
\rowcolor[HTML]{D9D9D9} 
35   & 4    & Dishwasher door opening                 & 7       & Door opening          \\
\rowcolor[HTML]{D9D9D9} 
36   & 4    & Drawer closing                          & 10      & Drawer closing        \\
\rowcolor[HTML]{D9D9D9} 
37   & 4    & Hammerhead shark pick and place         & 4       & 6D pick \& place        \\
\rowcolor[HTML]{D9D9D9} 
38   & 4    & Oil pouring                             & 5       & Pouring               \\
\rowcolor[HTML]{D9D9D9} 
39   & 4    & Almonds pouring                         & 6       & Pouring               \\
\rowcolor[HTML]{D9D9D9} 
40   & 4    & Chair pulling                           & 8       & Chair pulling         \\
\rowcolor[HTML]{D9D9D9} 
41   & 4    & Book pulling                            & 10      & Pulling from shelf    \\
\rowcolor[HTML]{D9D9D9} 
42   & 4    & Tissue pulling                          & 5       & Tissue pickup         \\
\rowcolor[HTML]{D9D9D9} 
43   & 4    & Paper bag pickup                        & 8       & Bag pickup            \\
44   & 5    & Microwave Door Opening                  & 7       & Door opening          \\
45   & 5    & Drawer closing                          & 10      & Drawer closing        \\
46   & 5    & Drawer opening                          & 10      & Drawer opening        \\
47   & 5    & Chair pulling                           & 10      & Chair pulling         \\
48   & 5    & Towel pulling from the fridge           & 7       & Towel pickup          \\
49   & 5    & DVD pulling                             & 10      & Pulling from shelf    \\
50   & 5    & Knob turning                            & 5       & Knob turning          \\
51   & 5    & Paper towel tube                        & 5       & Paper towel replacing \\
\rowcolor[HTML]{D9D9D9} 
52   & 6    & Door opening kitchen                    & 10      & Door opening          \\
\rowcolor[HTML]{D9D9D9} 
53   & 6    & Door opening bathroom                   & 7       & Door opening          \\
\rowcolor[HTML]{D9D9D9} 
54   & 6    & Drawer closing                          & 10      & Drawer closing        \\
\rowcolor[HTML]{D9D9D9} 
55   & 6    & Mini drawer closing                     & 10      & Drawer closing        \\
\rowcolor[HTML]{D9D9D9} 
56   & 6    & Dishwasher drawer opening               & 8       & Drawer opening        \\
\rowcolor[HTML]{D9D9D9} 
57   & 6    & Lantern pick and place                  & 9       & 6D pick \& place        \\
\rowcolor[HTML]{D9D9D9} 
58   & 6    & Chair pulling                           & 10      & Chair pulling         \\
\rowcolor[HTML]{D9D9D9} 
59   & 6    & Table pulling                           & 10      & Chair pulling         \\
\rowcolor[HTML]{D9D9D9} 
60   & 6    & Rag pull                                & 9       & Towel pickup          \\
\rowcolor[HTML]{D9D9D9} 
61   & 6    & Book pulling                            & 8       & Pulling from shelf    \\
\rowcolor[HTML]{D9D9D9} 
62   & 6    & Tissue pick up                          & 10      & Tissue pickup         \\
\rowcolor[HTML]{D9D9D9} 
63   & 6    & Bag pick up                             & 8       & Bag pickup            \\
\rowcolor[HTML]{D9D9D9} 
64   & 6    & Cushion lifting                         & 10      & Cushion flipping      \\
65   & 7    & Kitchen door closing                    & 10      & Door closing          \\
66   & 7    & Bathroom closet door opening            & 9       & Door opening          \\
67   & 7    & Drawer closing black wardrode           & 7       & Drawer closing        \\
68   & 7    & Drawer closing white wardrode           & 10      & Drawer closing        \\
69   & 7    & Drawer closing desk                     & 8       & Drawer closing        \\
70   & 7    & Drawer closing table                    & 8       & Drawer closing        \\
71   & 7    & Chair pulling                           & 9       & Chair pulling         \\
72   & 7    & Dining table chair pulling              & 5       & Chair pulling         \\
73   & 7    & Rag pulling                             & 8       & Towel pickup          \\
74   & 7    & Tissue paper pick up                    & 10      & Tissue pickup         \\
75   & 7    & Paper Towel pick up                     & 10      & Paper towel replacing \\
76   & 7    & Trash pickup                            & 8       & Bag pickup            \\
\rowcolor[HTML]{D9D9D9} 
77   & 8    & Door opening                            & 8       & Door opening          \\
\rowcolor[HTML]{D9D9D9} 
78   & 8    & Air fryer open                          & 9       & Air-fryer opening     \\
\rowcolor[HTML]{D9D9D9} 
79   & 8    & Air fryer close                         & 10      & Air-fryer closing     \\
\rowcolor[HTML]{D9D9D9} 
80   & 8    & Chair pulling                           & 10      & Chair pulling         \\
\rowcolor[HTML]{D9D9D9} 
81   & 8    & Unplugging                              & 6       & Unplugging            \\
\rowcolor[HTML]{D9D9D9} 
82   & 8    & Toilet rag pulling                      & 9       & Towel pickup          \\
\rowcolor[HTML]{D9D9D9} 
83   & 8    & Book pulling                            & 8       & Pulling from shelf    \\
\rowcolor[HTML]{D9D9D9} 
84   & 8    & Codenames pulling                       & 7       & Pulling from shelf    \\
\rowcolor[HTML]{D9D9D9} 
85   & 8    & Tissue pick up                          & 7       & Tissue pickup         \\
\rowcolor[HTML]{D9D9D9} 
86   & 8    & Paper towel roll pickup                 & 7       & Paper towel replacing \\
\rowcolor[HTML]{D9D9D9} 
87   & 8    & Food bag pick up                        & 8       & Bag pickup            \\
\rowcolor[HTML]{D9D9D9} 
88   & 8    & Cushion flip                            & 10      & Cushion flipping      \\
\rowcolor[HTML]{D9D9D9} 
89   & 8    & Toilet flushing                         & 9       & Random                \\
90   & 9    & Door closing                            & 10      & Door closing          \\
91   & 9    & Door opening                            & 7       & Door opening          \\
92   & 9    & Bathroom drawer closing                 & 10      & Drawer closing        \\
93   & 9    & Kitchen drawer closing                  & 10      & Drawer closing        \\
94   & 9    & Kitchen drawer opening                  & 6       & Drawer opening        \\
95   & 9    & Hat pickup                              & 9       & Misc object pickup    \\
96   & 9    & Chair pulling                           & 9       & Chair pulling         \\
97   & 9    & Light switch                            & 6       & Light switch          \\
98   & 9    & Rag pulling                             & 10      & Towel pickup          \\
99   & 9    & Book pulling                            & 7       & Pulling from shelf    \\
100  & 9    & Paper bag pick up                       & 10      & Bag pickup            \\
\rowcolor[HTML]{D9D9D9} 
101  & 10   & Door Closing                            & 10      & Door closing          \\
\rowcolor[HTML]{D9D9D9} 
102  & 10   & Drawer Closing                          & 10      & Drawer closing        \\
\rowcolor[HTML]{D9D9D9} 
103  & 10   & Air fryer opening                       & 10      & Air-fryer opening     \\
\rowcolor[HTML]{D9D9D9} 
104  & 10   & Air fryer closing                       & 10      & Air-fryer closing     \\
\rowcolor[HTML]{D9D9D9} 
105  & 10   & Light switch                            & 8       & Light switch          \\
\rowcolor[HTML]{D9D9D9} 
106  & 10   & Hand towel (rag) pulling                & 7       & Towel pickup          \\
\rowcolor[HTML]{D9D9D9} 
107  & 10   & Book pulling                            & 10      & Pulling from shelf    \\
\rowcolor[HTML]{D9D9D9} 
108  & 10   & Paper towel                             & 9       & Paper towel replacing \\
\rowcolor[HTML]{D9D9D9} 
109  & 10   & Cushion straightening                   & 10      & Cushion flipping     
\label{tab:all-tasks}
\end{longtable}
\begin{figure}[p!]
    \centering
    \vspace{-2em}
    \includegraphics[width=0.93\textwidth]{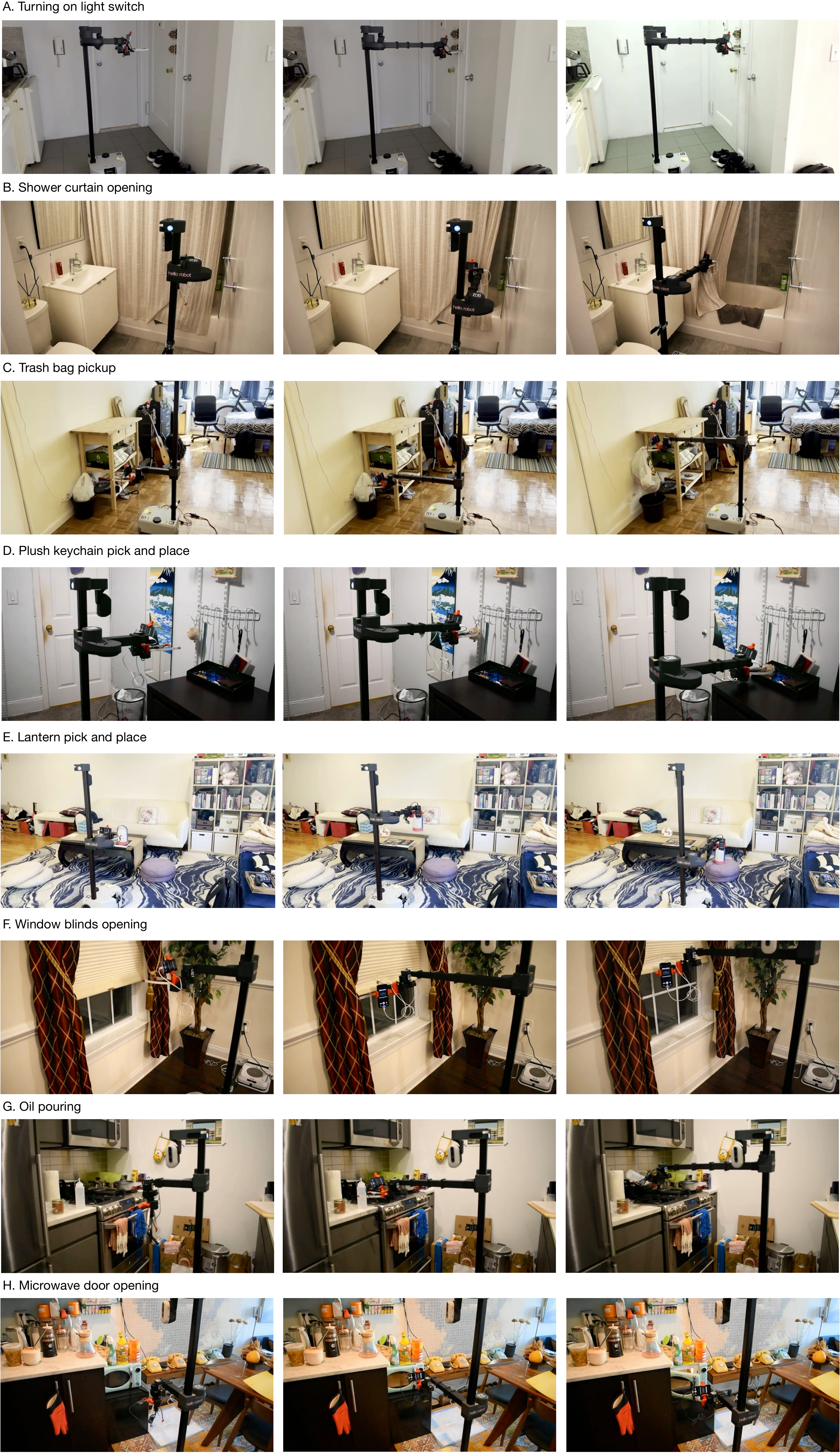}
    \caption{A small subset of 8 robot rollouts from the \numhomeexperiments{} tasks that we tried in homes. A complete set of rollout videos can also be found at our website: \url{\website /\#videos}}
    \label{fig:robot-run-sample}
\end{figure}

\begin{figure}[t]
    \centering
    \includegraphics[width=\linewidth]{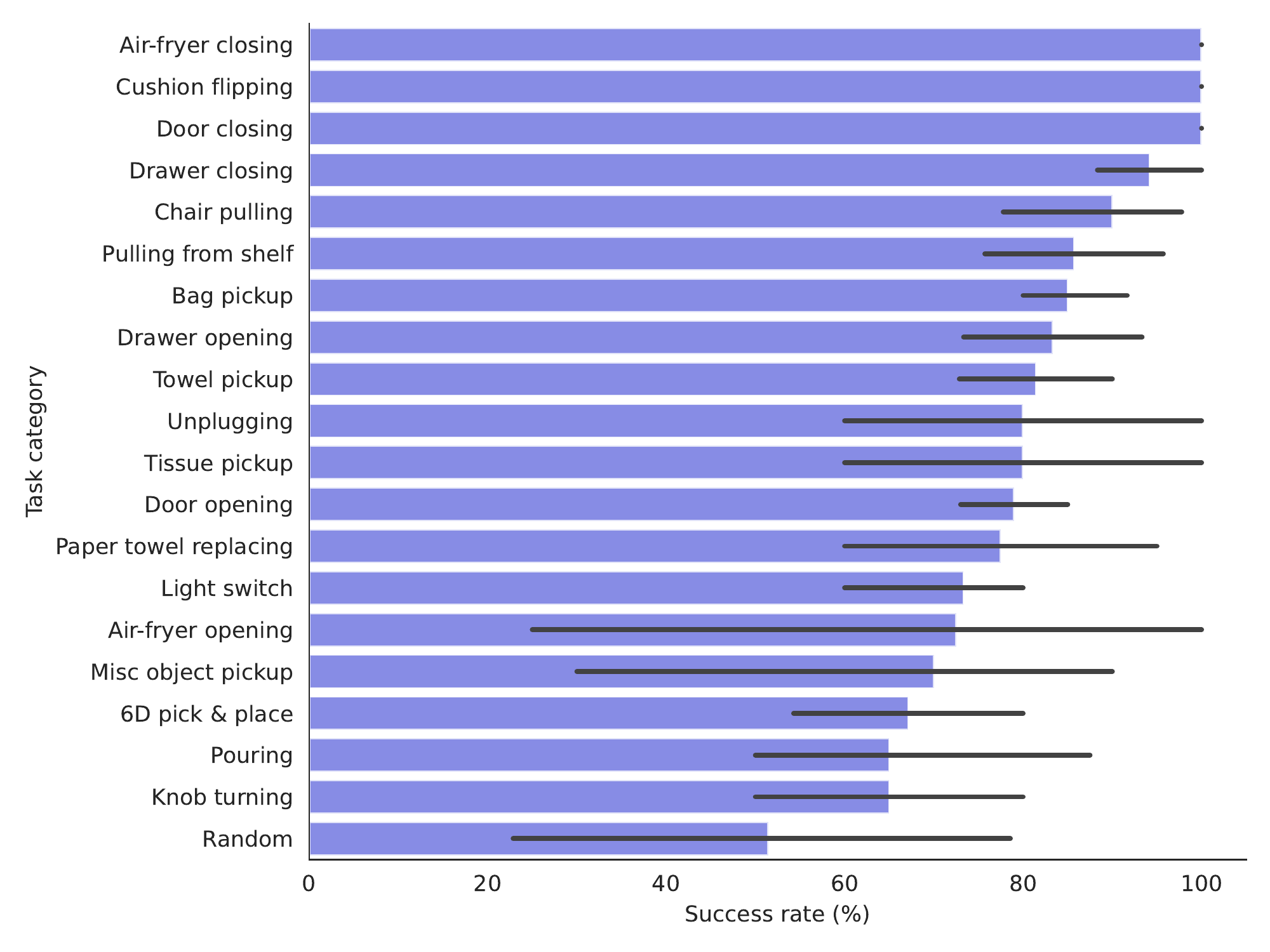}
    \caption{Success rate of our 20 different task groups, with the variance in each group's success rate shown in the error bar.}
    \label{fig:task-success-rates}
\end{figure}

\begin{figure}[t]
    \centering
    \includegraphics[width=0.65\linewidth]{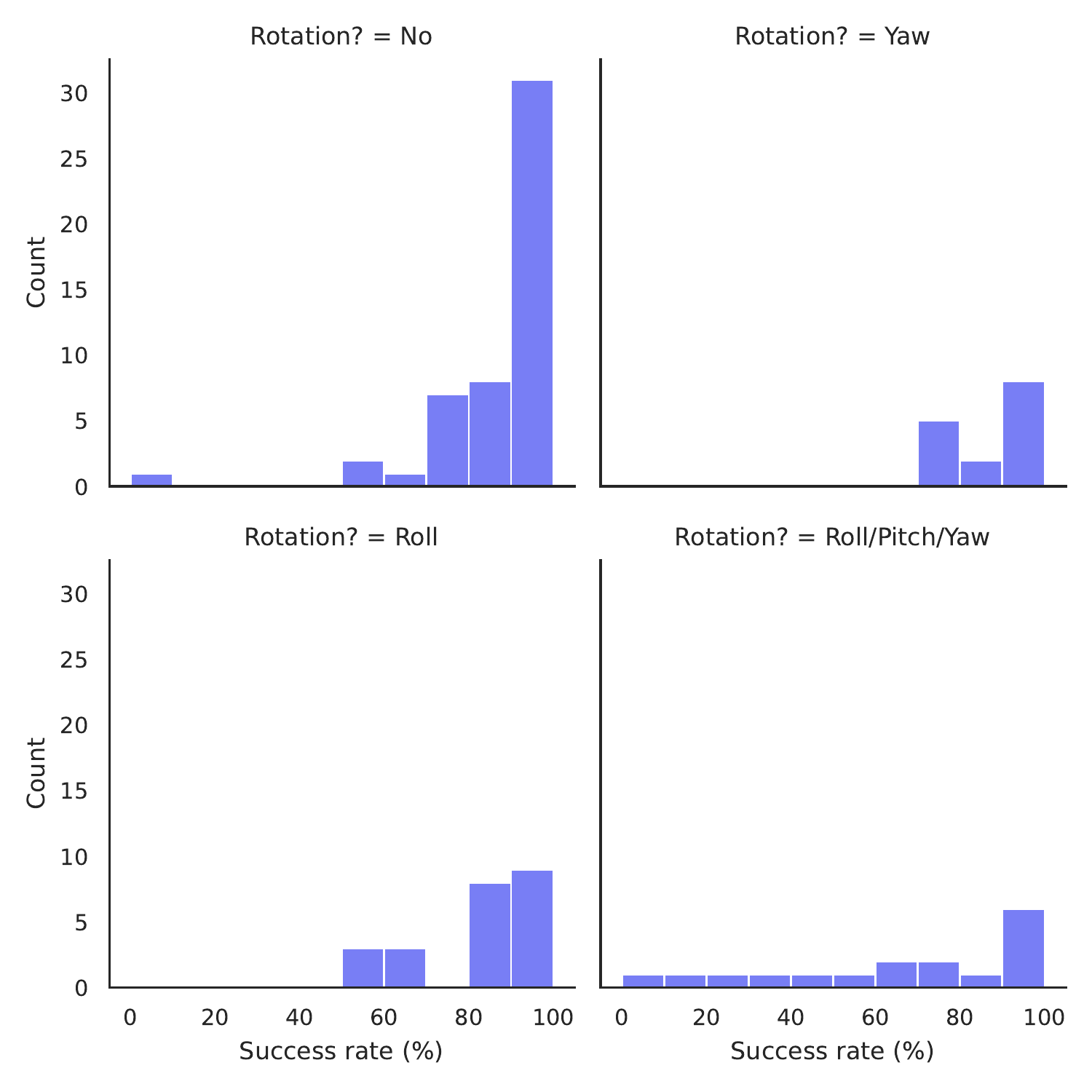}
    \caption{Success rate breakdown by type of actions needed to solve the task. The X-axis shows the number of successes out of 10 rollouts, and the Y-axis shows number of tasks with the corresponding number of success.}
    \label{fig:rotational-actions}
\end{figure}
\clearpage
\begin{figure}[t]
    \centering
    \includegraphics[width=\linewidth]{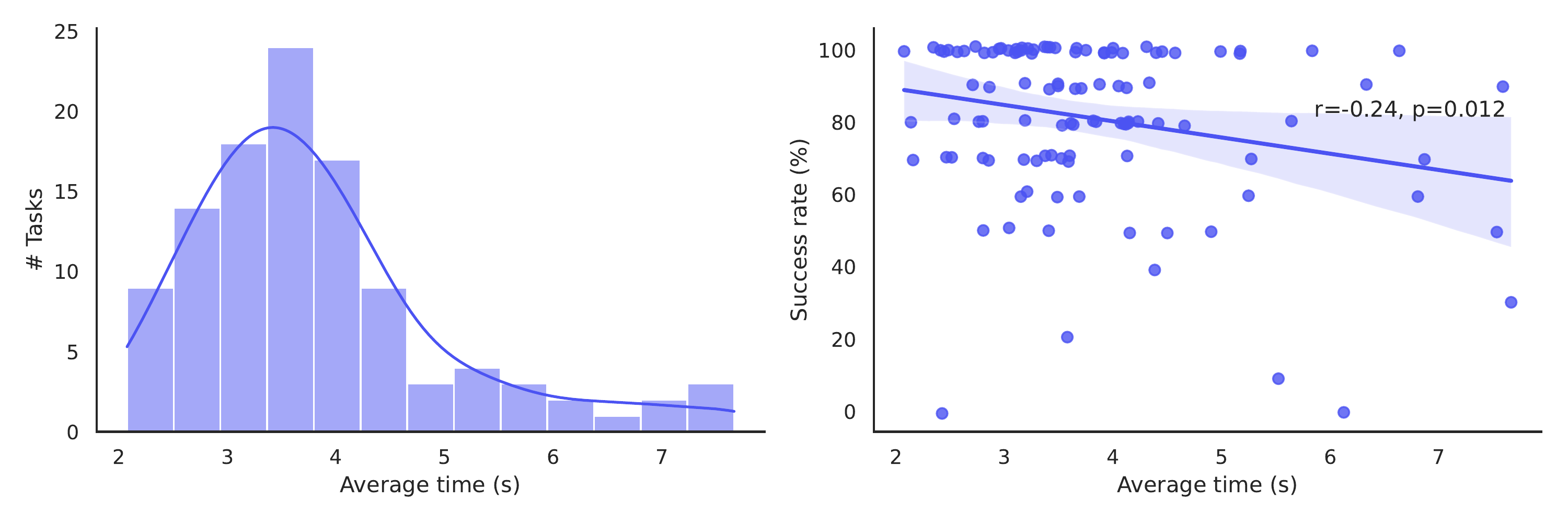}
    \caption{(a) Distribution of time (in seconds) taken to demonstrate a task on our experiment setup. The mean time taken to complete one demonstration is 3.82 seconds, and the median time taken is 3.49 seconds. 
    (b) Correlation analysis between time taken to demonstrate a task and the success rate of the associated robot policy.}
    \label{fig:correlation}
\end{figure}

\subsection{Understanding the Performance of \method{}}
\label{sec:understanding-performance}
On a broad level, we cluster our tasks into 20 broad categories, 19 task specific and one for the miscellaneous tasks. There are clear patterns in how easy or difficult different tasks may be, compared to each other.

\subsubsection{Breakdown by Task Type}
\label{sec:by-task-type}
We can see from Figure~\ref{fig:task-success-rates} that Air Fryer Closing and Cushion Flipping are the task groups with the highest average success rate (100\%) while the task group with the lowest success rate is 6D pick \& place (56\%). We found that 6D pick and place tasks generally fail because they generally require robot motion in a variety of axes: like translations and rotations at different axes at different parts of the trajectory, and we believe more data may alleviate the issue. We discuss the failure cases further in Section~\ref{sec:failures}.

\subsubsection{Breakdown by Action Type}
\label{sec:by-action-tyoe}
We can cluster the tasks into buckets by their difficulty as shown in Figure~\ref{fig:rotational-actions}. We find that the type of movement affects the success rate of the tasks. Specifically, the distribution of success rates for tasks which do not require any wrist rotation is skewed much more positively compared to tasks where we need either yaw or roll, or a combination of yaw, pitch, and roll. Moreover, the distribution of successes for tasks which require 6D motion is the flattest, which shows that tasks requiring full 6D motions are harder compared to tasks where \method{} doesn't require full 6D motion.

\subsubsection{Correlation between demo time and difficulty}
\label{sec:correlation-analysis}

Here, we try to analyze the relationship between the difficulty of a task group when done by the robot, and the time required to complete the task by a human. To understand the relationship between these two variables related to a task, we perform a regression analysis between them.

We see from Figure~\ref{fig:correlation} that there is a weak negative correlation ($r = -0.24$, with $p = 0.012 < 0.05$) between the amount of time taken to complete a demo by the human demonstrator and how successful the robot is at completing the task. This analysis implies that while longer tasks may be harder for the robot to accomplish, there are other factors that contribute to making a task easy or difficult.

\begin{figure}[t!]
    \centering
    \includegraphics[width=\linewidth]{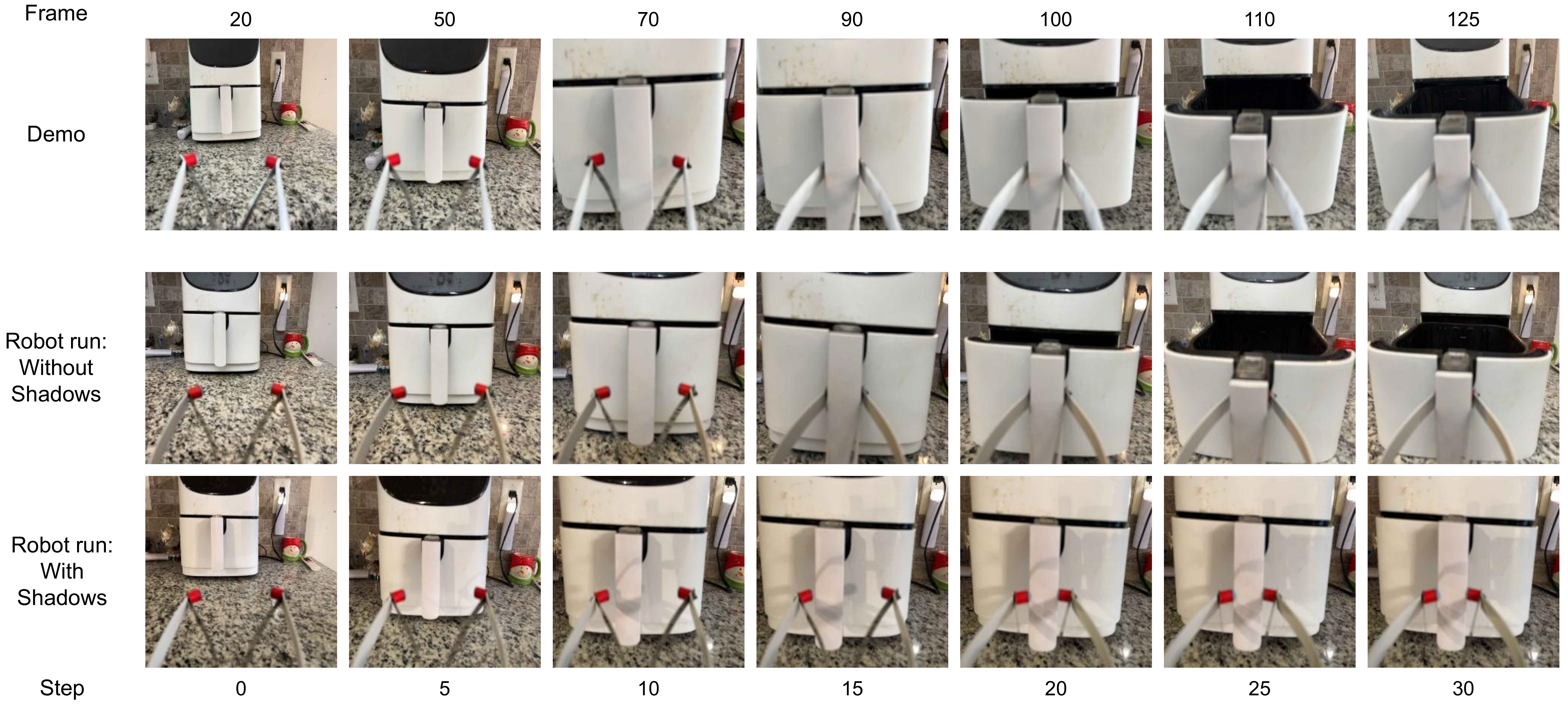}
    \caption{First-person POV rollouts of Home 1 Air Fryer Opening comparing (top row) the original demonstration environment, against robot performance in environments with (middle row) similar lighting, and (bottom row) altered lighting conditions with additional shadows.}
    \label{fig:failure-1}
\end{figure}
\begin{figure}[t!]
    \centering
    \includegraphics[width=\linewidth]{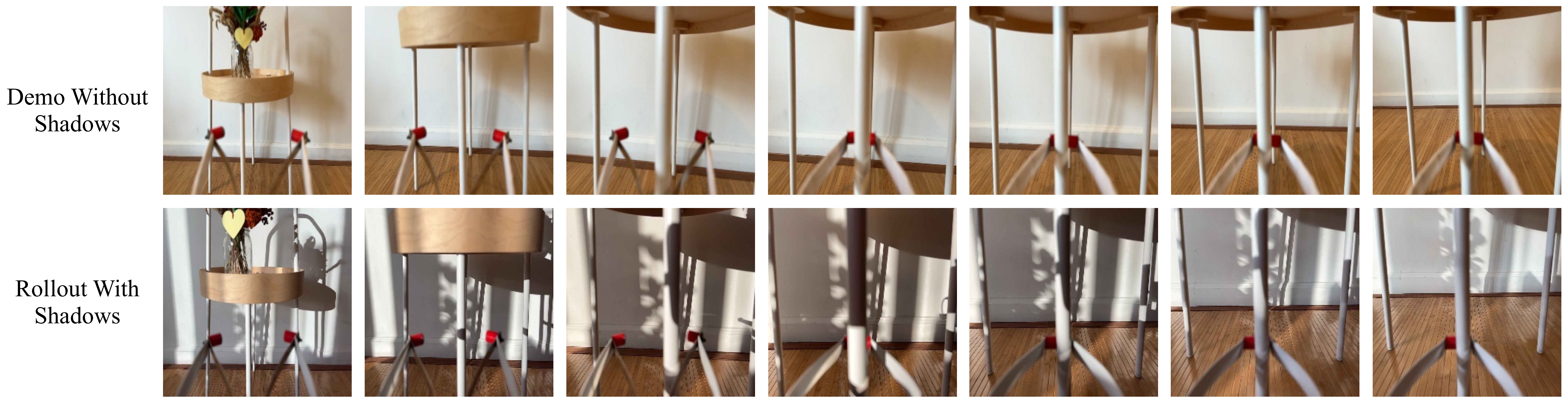}
    \caption{First person view from the iPhone from the (top row) \stick{} during demonstration collection and (bottom row) the robot camera during rollout. Even with strong shadows during rollout, the policy succeeds in pulling the table.}
    \label{fig:success-1}
\end{figure}
\begin{figure}[t!]
    \centering
    \includegraphics[width=\linewidth]{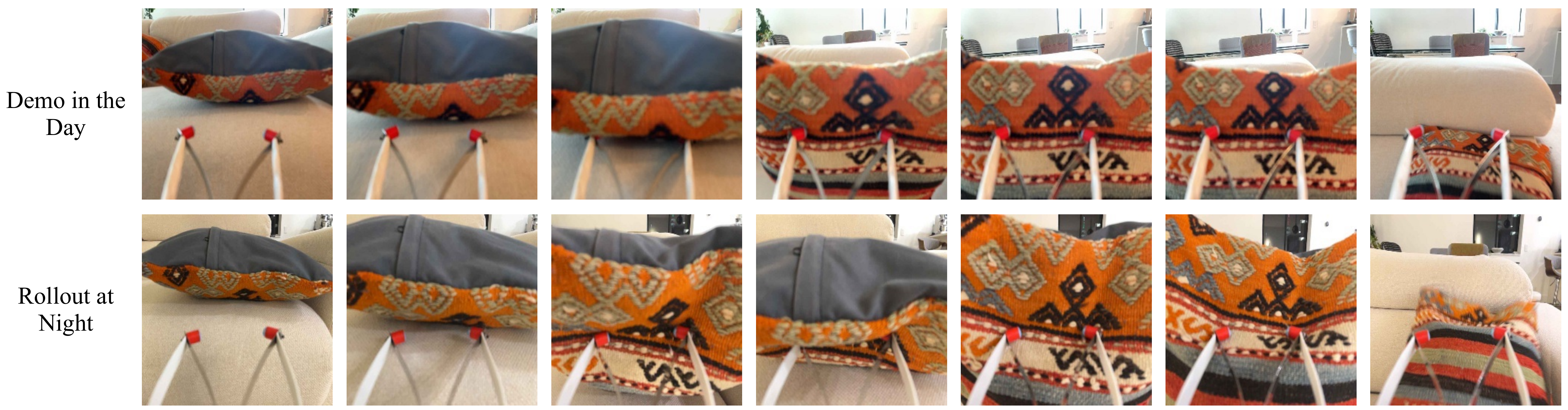}
    \caption{First person view from the iPhone from the (top row) \stick{} during demo collection and (bottom row) robot camera during rollout. The demonstrations were collected during early afternoon while rollouts happened at night; but because of the iPhone's low light photography capabilities, the robot view is similar.}
    \label{fig:success-2}
\end{figure}

\subsection{Failure Modes and Analysis}
\label{sec:failures}
\subsubsection{Lighting and shadows}

In many cases, the demos were collected in different lighting conditions than the policy execution. Generally, with enough ambient lighting, our policies succeeded regardless of day and night conditions. However, we found that if there was a strong shadow across the task space during execution that was not there during data collection, the policy may behave erratically.

The primary example of this is from Home 1 Air Fryer Opening (see Figure~\ref{fig:failure-1}), where the strong shadow of the robot arm caused our policy to fail. Once we turned on an overhead light for even lighting, there were no more failures. However, this shadow issue is not consistent, as we can see in Figure~\ref{fig:success-1}, where the robot performs the Home 6 table pulling task successfully despite strong shadows.

In many cases with lighting variations, the low-light photography capabilities of the iPhone helped us generalize across lighting conditions. For example, in Home 8 cushion straightening (Figure~\ref{fig:success-2}), we collected demos during the day and ran the robot during the night. However, from the robot perspective the difference in light levels is negligible.

\subsubsection{Sensor limitations}
\label{sec:failures:sensor-limits}
\begin{figure}[t]
    \vspace{0.5em}
    \centering
    \includegraphics[width=\linewidth]{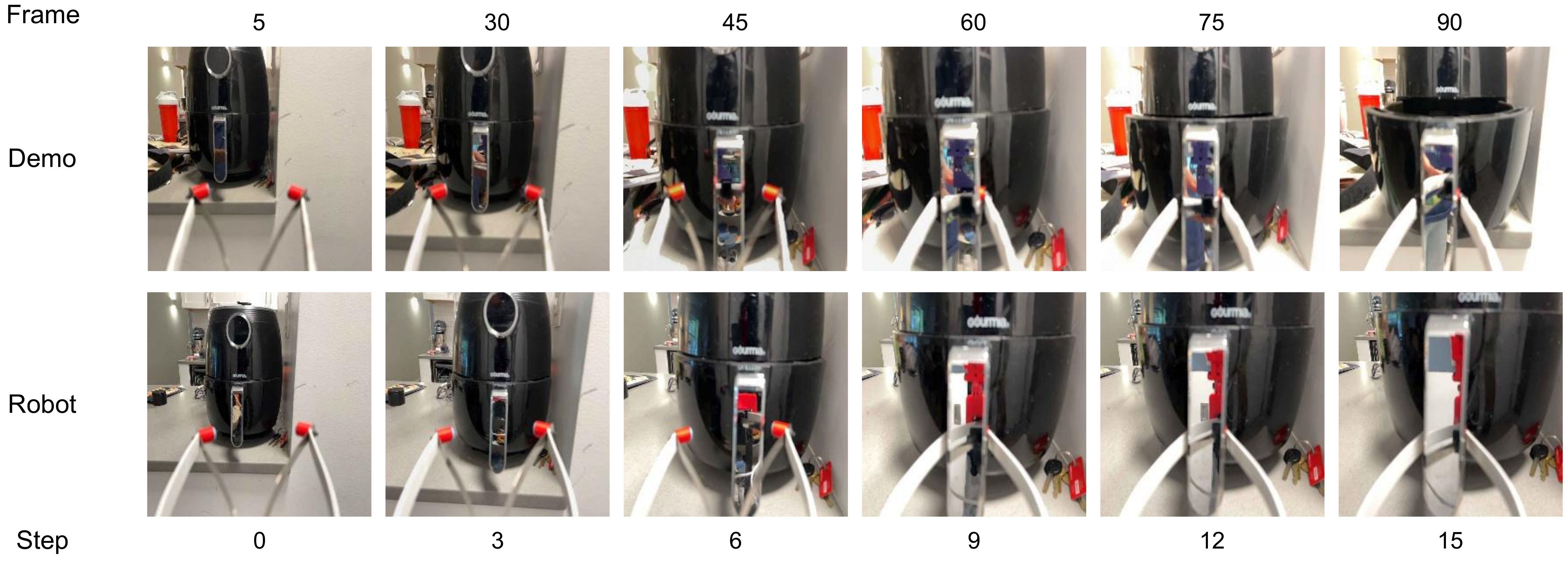}
    \caption{First-person POV rollouts of Home 3 Air Fryer Opening showcasing (top row) a demonstration of the task and (bottom row) robot execution.}
    \label{fig:failure-2}
    \vspace{-0.5em}
\end{figure}
\begin{figure}[t]
    \vspace{1.5em}
    \centering
    \includegraphics[width=\linewidth]{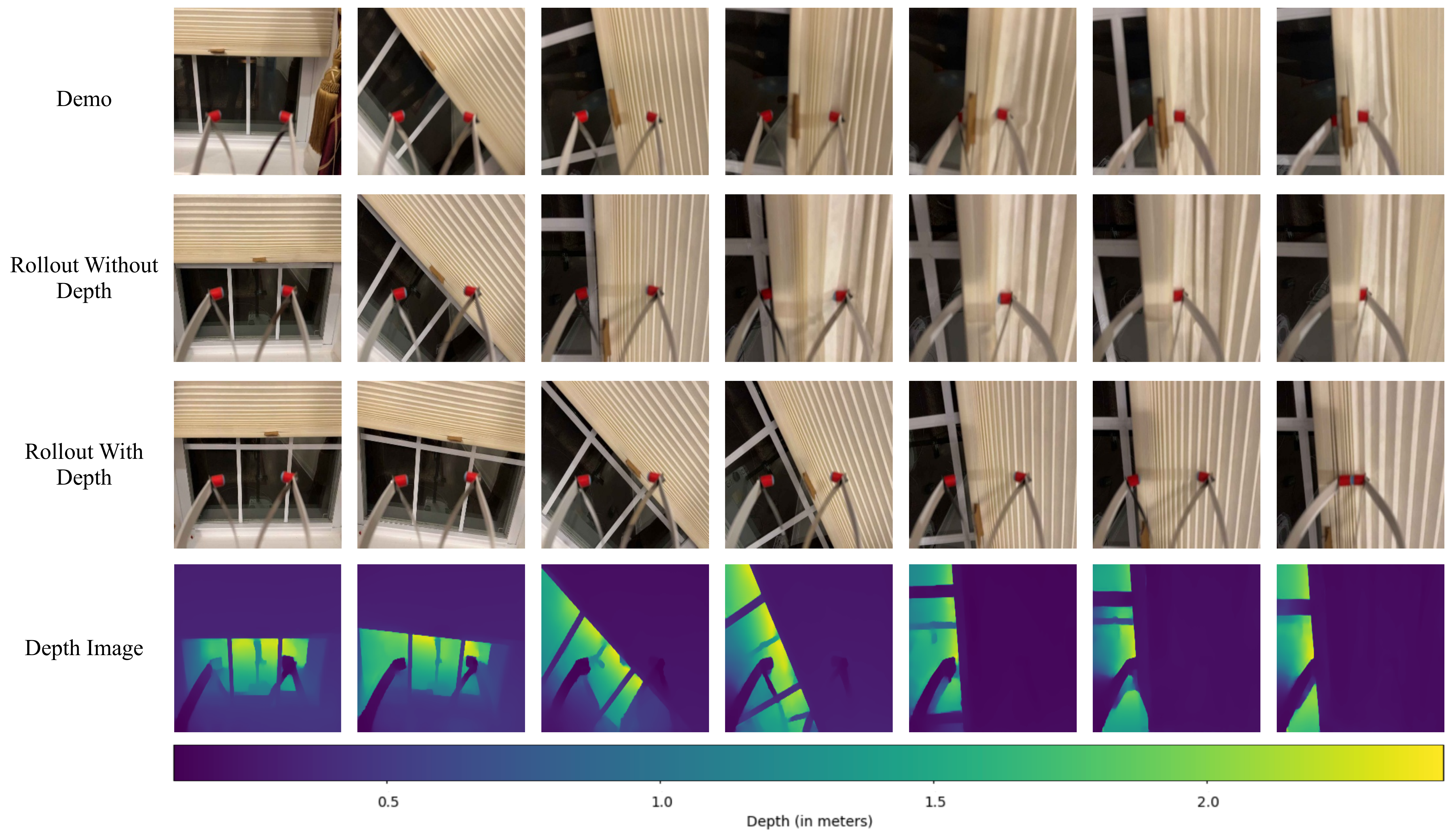}
    \caption{Opening an outward facing window blind (top row) both without depth (second row) and with depth (third row). The depth values (bottom row) for objects outside the window are high noisy, which cause the depth-aware behavior model to go out of distribution.}
    \label{fig:failure-3}
    \vspace{-1.5em}
\end{figure}

One of the limitations of our system is that we use a lidar-based depth sensor on the iPhone. Lidar systems are generally brittle at detecting and capturing the depth of shiny and reflective objects. As a result, around reflective surfaces we may get a lot of out-of-distribution values on our depth channel and our policies can struggle.

A secondary problem with reflective surfaces like mirrors is that we collect demonstrations using the \stick{} but run the trained policies on the robot. In front of a mirror, the demonstration may actually end up recording the demo collector in the mirror. Then, once the policy is executed on the robot, the reflection on the mirror captures the robot instead of the demonstrator, and so the policy goes out-of-distribution and fails.

One of the primary examples of this is Home 3 Air Fryer Opening (Figure~\ref{fig:failure-2}). There, the air fryer handle was shiny, and so had both bad depth and captured the demonstration collector reflection which was different from the robot reflection. As a result, we had 0/10 successes on this task.

Another example is Home 1 vertical window blinds opening, where the camera faced outwards in the dark and provided many out-of-distribution values for the depth (Figure~\ref{fig:failure-3}). In this task, depth-free models performed better (10/10 successes) than depth-using models (2/10 successes) because of such values.

\subsubsection{Robot hardware limitations}
\label{sec:failures:hardware-limits}
\begin{figure}[t]
    \centering
    \includegraphics[width=\linewidth]{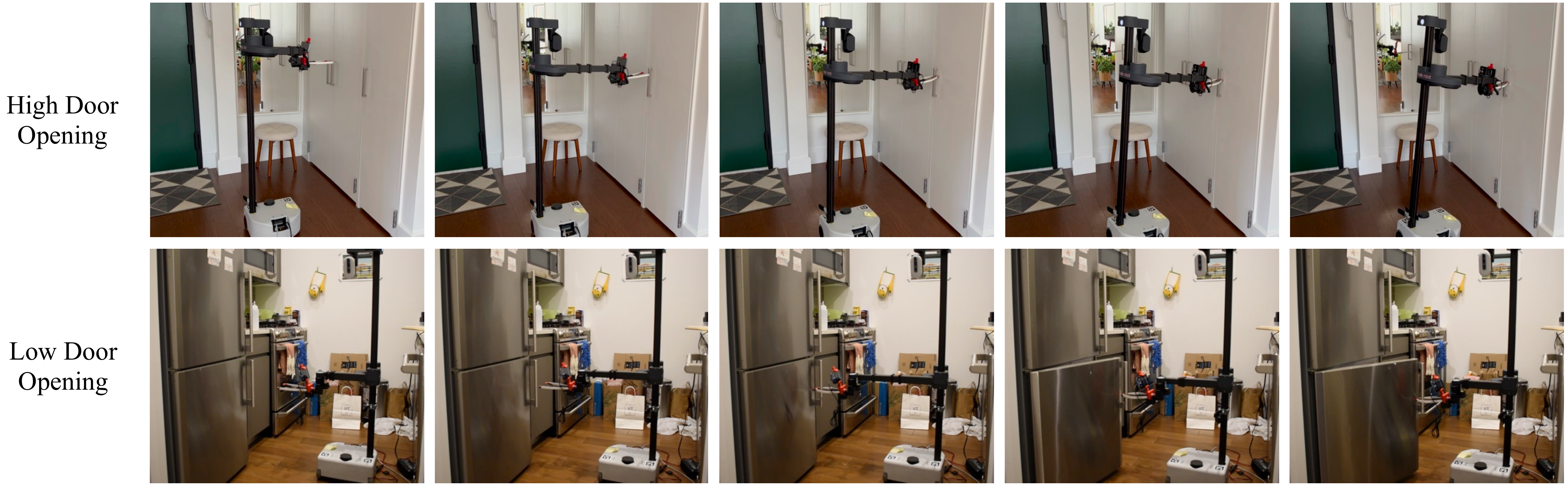}
    \caption{The robot pulling on a heavy door handle (top row) high up from the ground and (bottom row) closer the ground. Since the robot is bottom heavy, the first case starts tipping the robot while the second case succeeds.}
    \label{fig:failure-4}
\end{figure}

Our robot platform, Hello Robot Stretch RE1, was robust enough that we were able to run all the home experiments on a single robot with only minor repairs. However, there are certain hardware limitations that caused several of our tasks to fail.

The primary constraint we faced was the robot’s height limit. While the Stretch is tall, the manipulation space caps out at 1m, and thus a lot of tasks like light switch flicking or picking and placing from a high position are hard for the robot to do. Another challenge with the robot is that since the robot is tall and bottom-heavy, putting a lot of pulling or pushing force with the arm near the top of the robot would tilt the robot rather than moving the arm (Figure~\ref{fig:failure-4}), which was discussed in~\cite{kemp2021design}. Comparatively, the robot was much more successful at opening heavy doors and pulling heavy objects when they were closer to the ground than not, as shown in the same figure. A study of such comparative pulling forces needed can be found in~\cite{jain2010complex, jain2013improving}.

Knob turning, another low performing task, had 65\% success rate because of the fine manipulation required: if the robot’s grasp is not perfectly centered on the knob, the robot may easily move the wrist without moving the knob properly.

\subsubsection{Temporal dependencies}
\label{sec:failures:temporal-dependencies}

\begin{figure}[t]
    \centering
    \includegraphics[width=\linewidth]{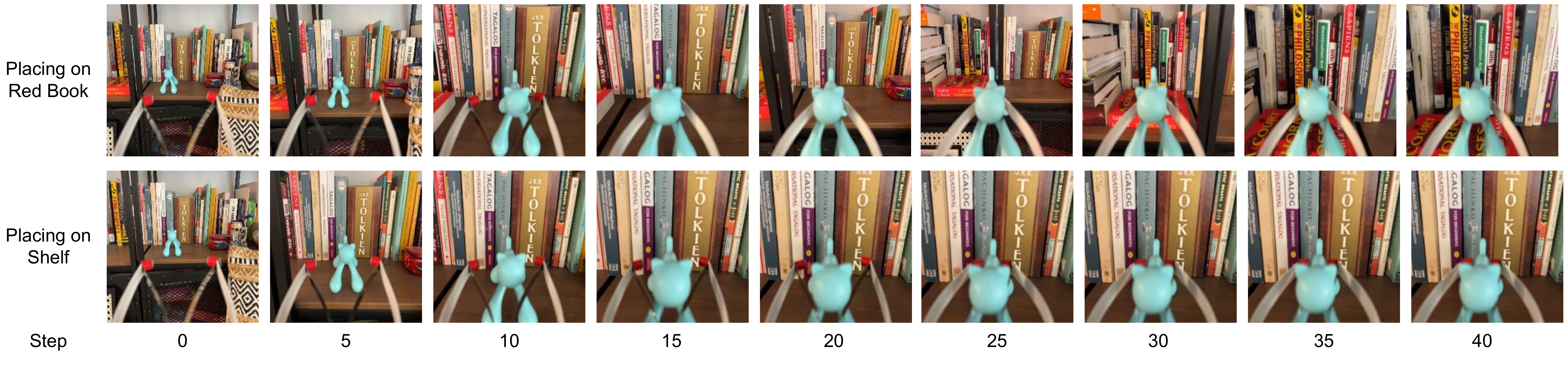}
    \caption{First-person POV rollouts of Home 3 Pick and Place comparing (top) a policy trained on demos where the object is picked and placed onto a red book on a different shelf and (bottom) a policy trained on demos where the object is picked and placed onto that same shelf without a red book. In the second case, since there is no clear signal for when to place the object, the BC policy keeps moving left and fails to complete the task.}
    \label{fig:failure-5}
\end{figure}

Finally, while our policy only relies on the last observations, for a lot of tasks, being able to consider temporal dependency would give us a much more capable policy class. For example, for a lot of Pick and Place tasks, the camera view right after picking up an object and the view right before placing the object may look the same. In that case, a policy that is not aware of time or previous observations gets confused and can’t decide between moving forward and moving backwards. A clear example of this is in Home 3 Pick and Place onto shelf (Figure~\ref{fig:failure-5}), where the policy is not able to place the object if the pick location and the place location (two shelf racks) look exactly the same, resulting in 0/10 successes. However, if the policy is trained to pick and place the exact same object on a different surface (here, a red book on the shelf rack), the model succeeds 7/10 times. A policy with temporal knowledge~\cite{brohan2022rt1, chi2023diffusion, shafiullah2022behavior} could solve this issue.

\clearpage
\subsection{Ablations}
\label{sec:ablations}
We created a benchmark set of tasks in our lab, with a setup that closely resembles a home, to be able to easily run a set of ablation experiments for our framework. To compare various parts of our system, we compare them with alternate choices, and show the relative performance in different tasks. These ablation experiments evaluate different components of our system and how they contribute to our performance. The primary elements of our model that we ran ablations over are the visual representation, number of demonstrations required for our tasks, depth perception, expertise of the demonstrator, and the need for a parametric policy.

\subsubsection{Alternate visual representation models}
\label{sec:ablations:representation}
\begin{figure}[t]
\centering
\begin{subfigure}{.5\textwidth}
    \centering
    \includegraphics[width=\linewidth]{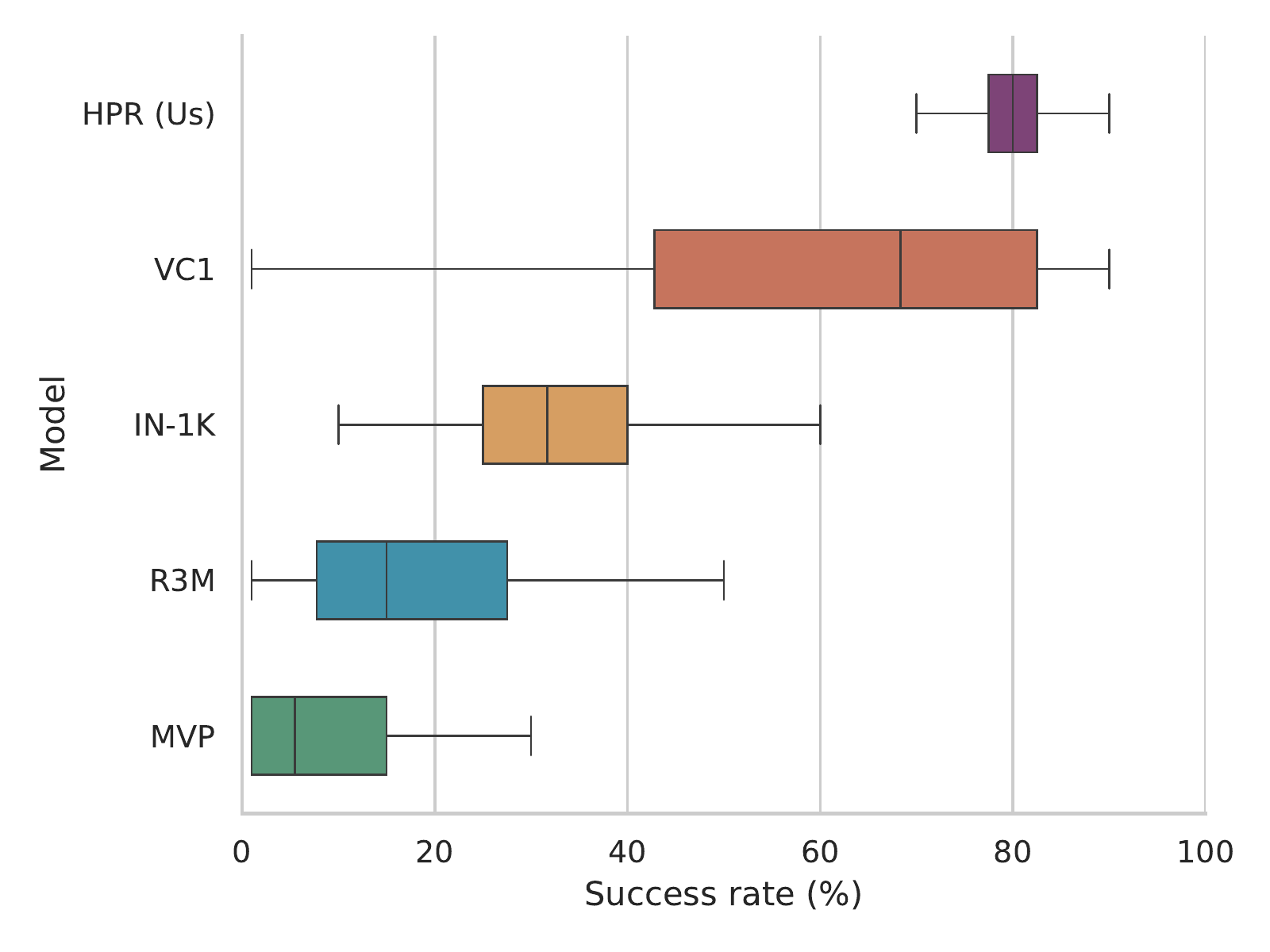}
    \caption{Lab tasks}
    \label{fig:sub1}
\end{subfigure}%
\begin{subfigure}{.5\textwidth}
    \centering
    \includegraphics[width=\linewidth]{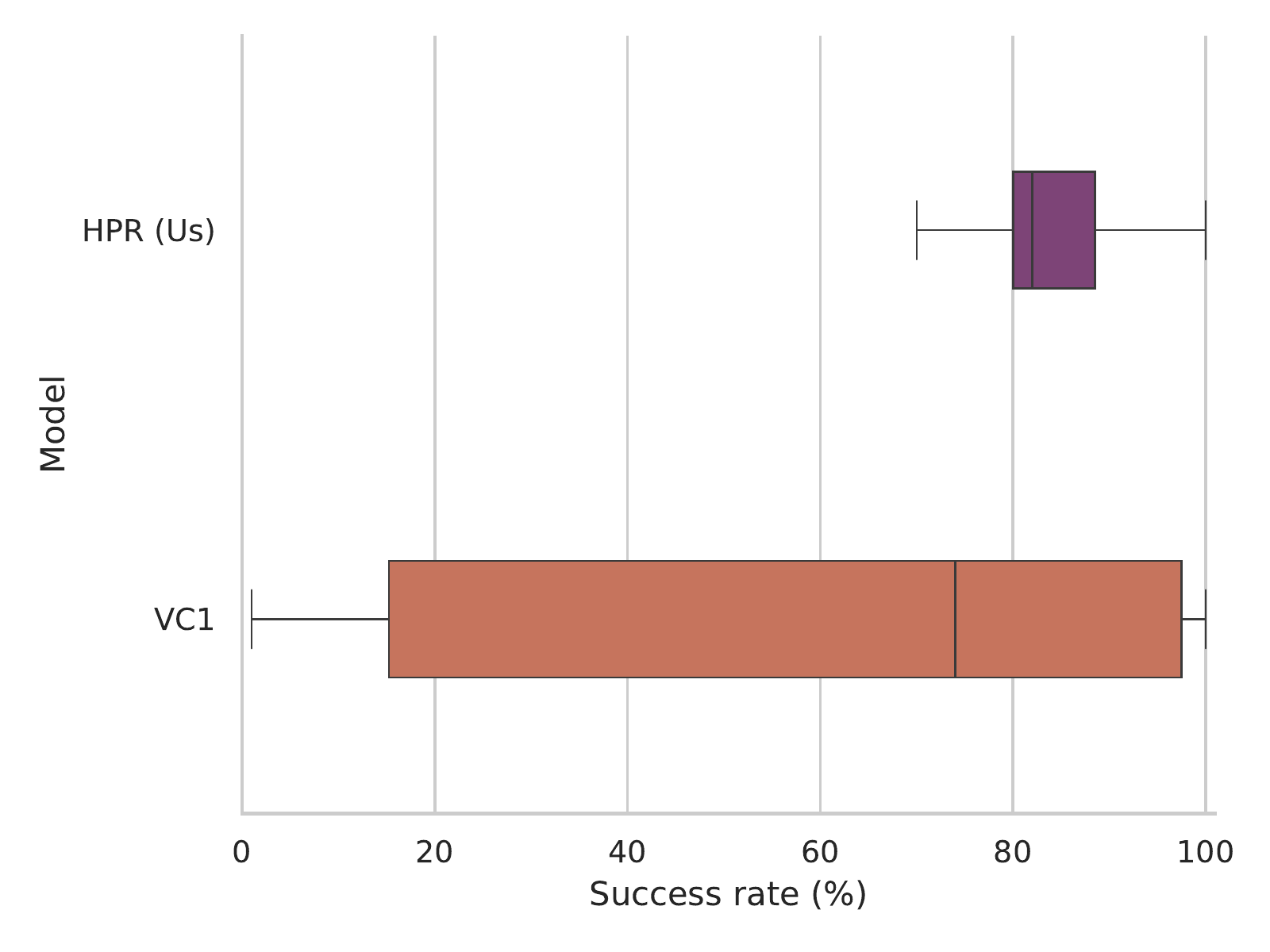}
    \caption{Home tasks}
    \label{fig:sub2}
\end{subfigure}
\caption{Comparison between different representation models at a set of tasks done in (a) our lab and (b) in a real home enviroment. As we can see, VC-1 is the representation model closest to ours in performance, however it has a high variance behavior where it either performs well or fails to complete the task entirely. The X-axis shows task completion rate distribution with the error bars showing the 95\% confidence interval.}
\label{fig:test}
\end{figure}

Our alternate visual representation comparison is with other pretrained representation models such as MVP~\cite{xiao2022mvp}, R3M~\cite{nair2022r3m}, VC1~\cite{majumdar2023vc1}, and a pretrained ImageNet-1k~\cite{he2016deep, deng2009imagenet} model. We compare them against our own pretrained models on the benchmark tasks, and compare the performances.

We see that in our benchmark environments, VC1 is the only representation that comes close to our trained representation. As a result, we ran some more experiments with VC1 representation in a household environment. As we can see, while VC1 is closer in performance to our model compared to IN-1K, R3M and MVP, it under-performs our model in household environments. However, VC-1 shows an interesting pattern of bimodal behavior: in each enviroment it either performs comparatively to HPR, or fails to complete the task entirely.

\begin{figure}[t]
    \centering
    \includegraphics[width=0.6\linewidth]{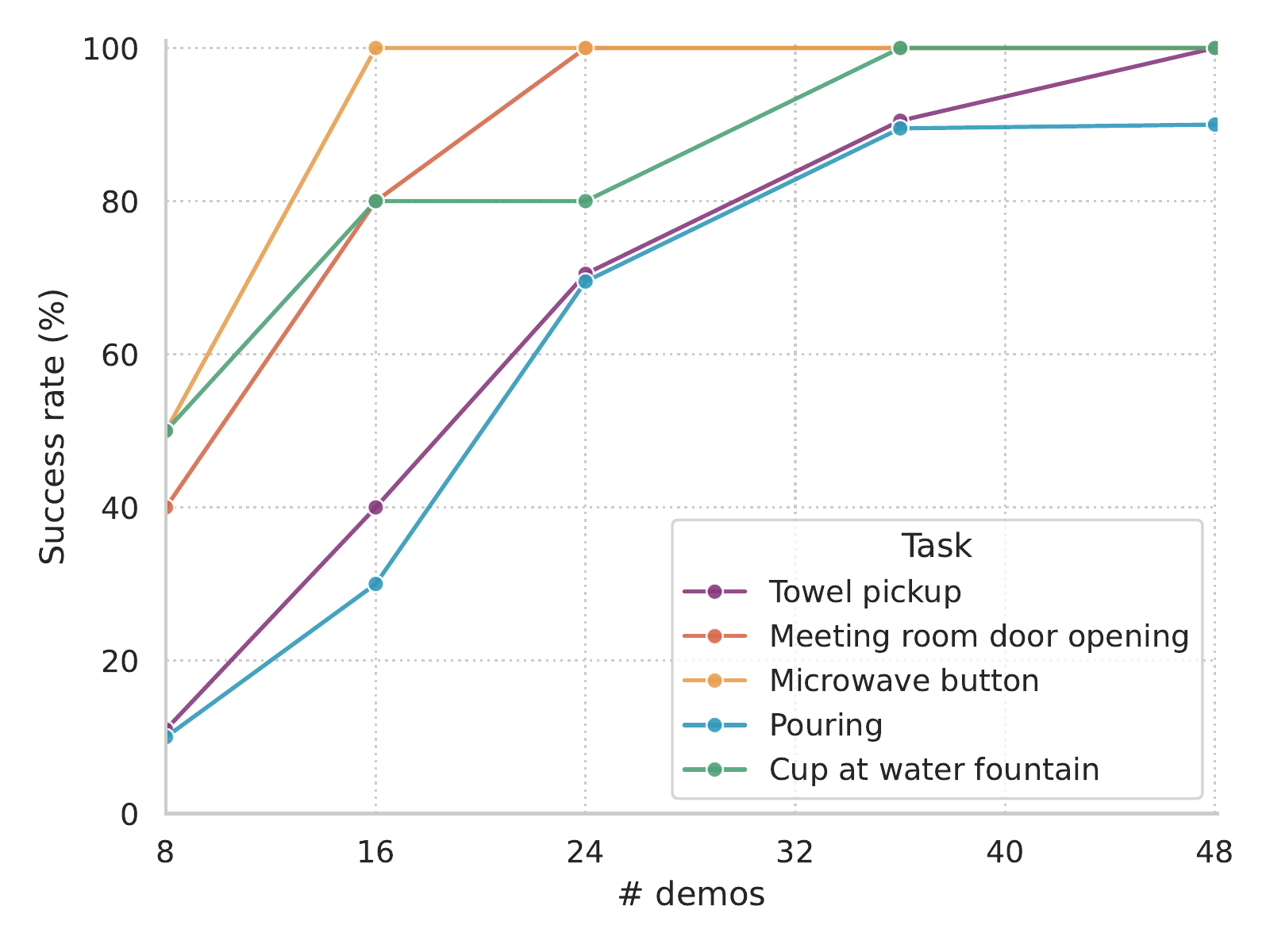}
    \caption{Success rates for a given number of demonstrations for five different tasks. We see how the success rate converges as the number of demonstrations increase. 
    }
    \label{fig:success-vs-num-demos}
\end{figure}

\subsubsection{Number of demonstrations required for tasks}
While we perform all our tasks with 24 demonstrations each, different tasks may require different numbers of demonstrations. In this set of experiments, we show how models trained on different numbers of demonstrations compare to each other.

As we see in Figure~\ref{fig:success-vs-num-demos}, adding more demonstrations always improves the performance of our system. Moreover, we see that the performance of the model scales with the number of demonstrations until it saturates. This shows us that on the average case, if our model can somewhat solve a task, we can improve the performance of the system by simply adding more demonstrations.

\begin{figure}[t]
    \centering
    \includegraphics[width=\linewidth]{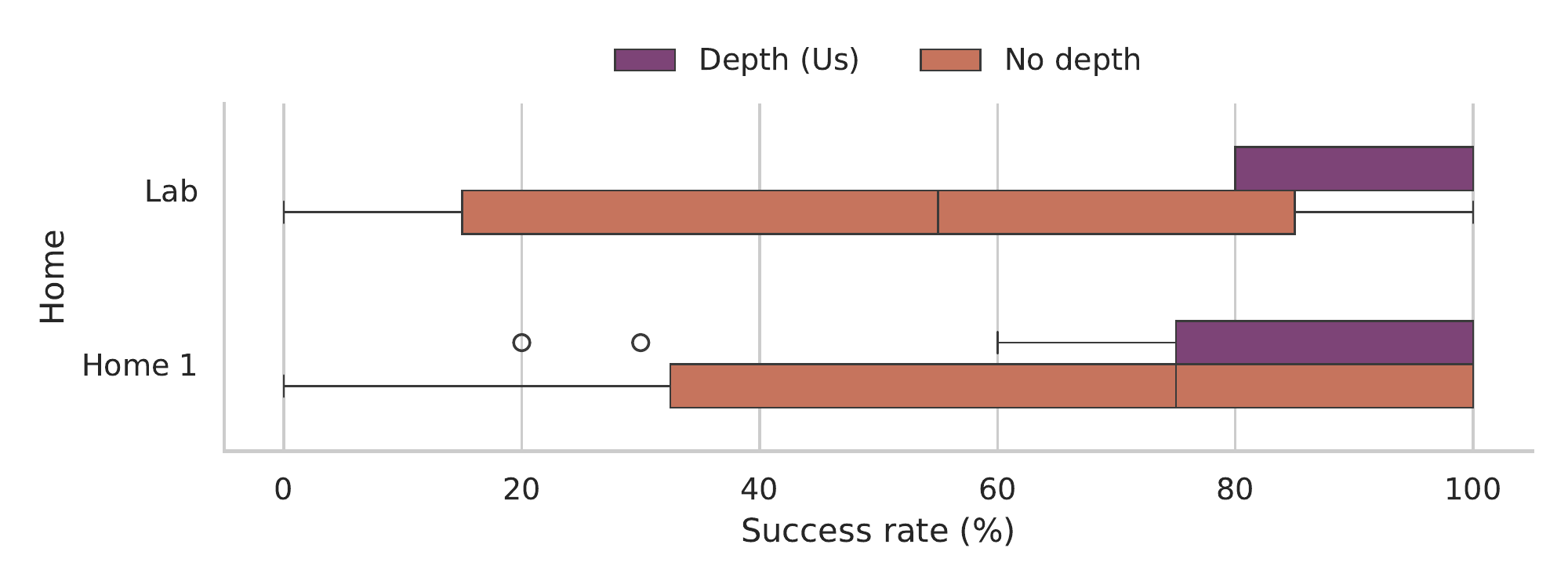}
    \caption{Barplot showing the distribution of task success rates in our two setups, one using depth and another not using depth. In most settings, using depth outperforms not using depth. However, there are some exceptional cases which are discussed in Section~\ref{sec:failures:sensor-limits}.}
    \label{fig:depth}
\end{figure}

\subsubsection{Depth Perception}
In this work, we use depth information from the iPhone to give our model approximate knowledge of the 3D structure of the world. Comparing the models trained with and without depth in Figure~\ref{fig:depth}, we can see that adding depth perception to the model helps it perform much better than the model with RGB-only input.

The failure modes for tasks without depth are generally concentrated around cases where the robot end-effector (and thus the camera) is very close to some featureless task object, for example a door or a drawer. Because such scenes do not have many features, it is hard for a purely visual imitation model without any depth information to know when exactly to close the gripper. On the other hand, the depth model can judge by the distance between the camera and the task surface when to open or close the gripper.

\subsubsection{Demonstrator Expertise}
Over the course of our project, we gained experience of how to collect demonstrations with the Stick. A question still remains of how much expertise is needed to operate the Stick and collect workable demonstrations with it.

For this experiment, we have two novice demonstrators collect demonstrations for two tasks in our lab setup. In Task 1, our collected data gave 100\% success, while in Task 2, our collected data gave 70\% success. Novice collector 1 collected data for Task 1 first and Task 2 second, while collector 2 collected data for Task 2 first and Task 1 second. Collector 1's data had 10\% success rate on Task 1, but had 70\% success on Task 2. Collector 2's data had 0\%  success on Task 2 but 90\% success on Task 1. From the data, we can see that while it may not be trivial initially to collect demonstrations and teach the robot new skills, with some practice both of our demonstrators were able to collect demonstrations that were sufficient.

\begin{figure}[t]
    \centering
    \includegraphics[width=\linewidth]{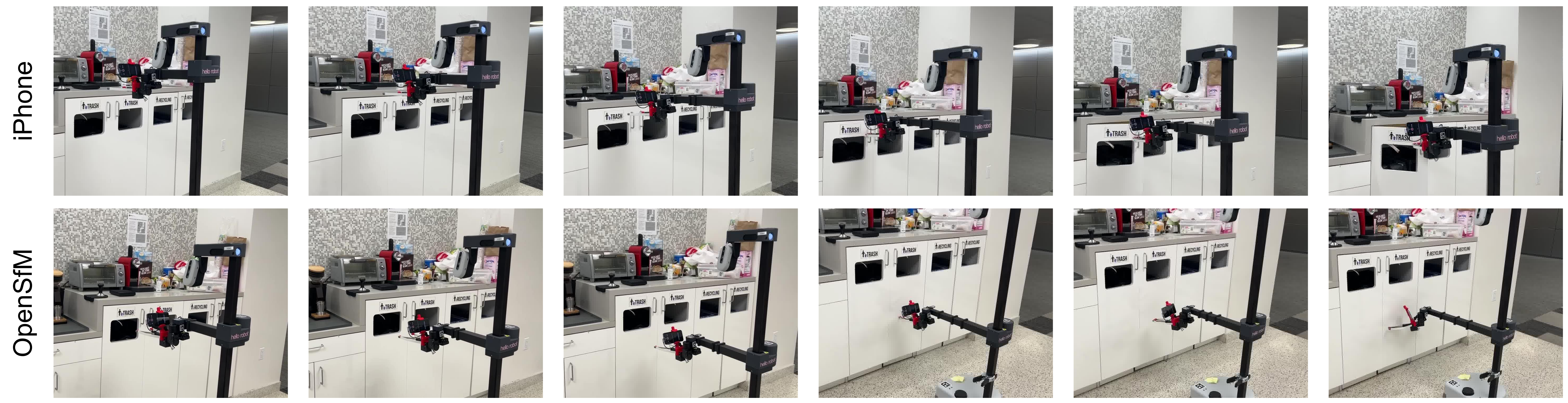}
    \caption{Open-loop rollouts from our demonstrations where the robot actions were extracted using (a) the odometry from iPhone and (b) OpenSfM respectively.}
    \label{fig:odometry}
\end{figure}

\subsubsection{Odometry}

In our system, we used the Stick odometry information based on the iPhone's odometry estimate.
Previous demonstration collection systems in works like~\cite{young2021playful, pari2021surprising} used structure-from-motion based visual odometry methods instead, like COLMAP~\cite{schonberger2016structure} and OpenSfM~\cite{adorjan2016opensfm}. In this section, we show the difference between the iPhone’s hardware-based and OpenSfM’s visual odometry methods, and compare the quality of the actions extracted from them.

As we can see from the Figure~\ref{fig:odometry}, OpenSfM-extracted actions are generally okay while the camera is far away from everything. However, it fails as soon as the camera gets very close to any surface and loses all visual features. The hardware odometry from the iPhone is much more robust, and thus the actions extracted from it are also reliable regardless of the camera view.
\section{Open Problems and Request for Research}
\label{sec:open-problems}

In this work we have presented an approach to scalable imitation learning that can be applied in household settings. However, there remains open problems that we must address before truly being able to bring robots to homes.

\begin{figure}[t]
    \centering
    \includegraphics[width=\textwidth]{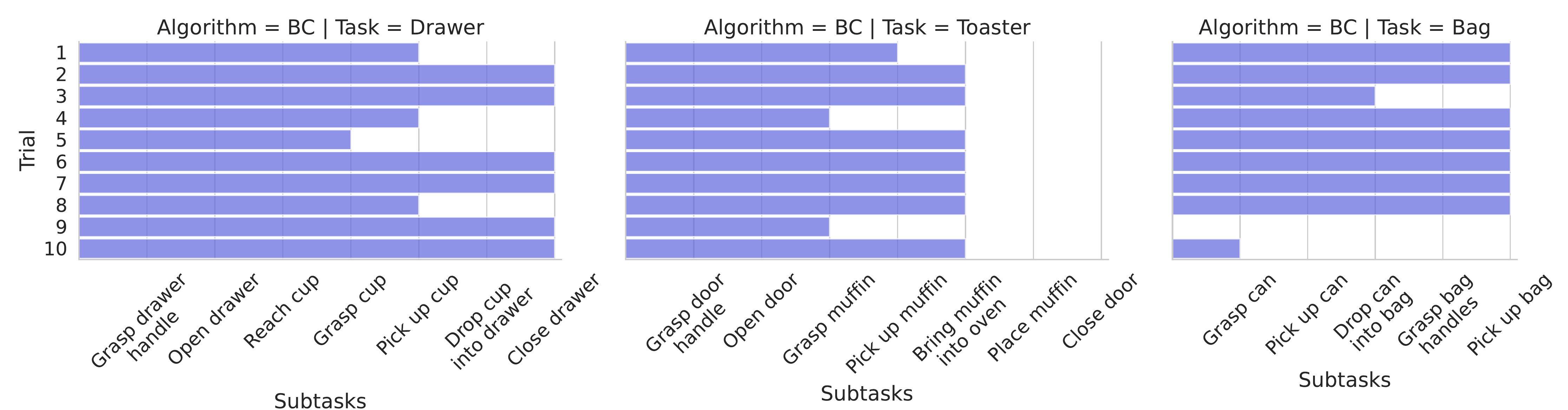}
    \caption{Analysis of our long-horizon tasks by subtasks. We see that \method{} can chain subtasks, although the errors can accumulate and make overall task success rate low.
    }
    \label{fig:long-horizon-analysis}
\end{figure}

\begin{figure}[p]
    \begin{subfigure}{\textwidth}
        \centering
        \includegraphics[width=\linewidth]{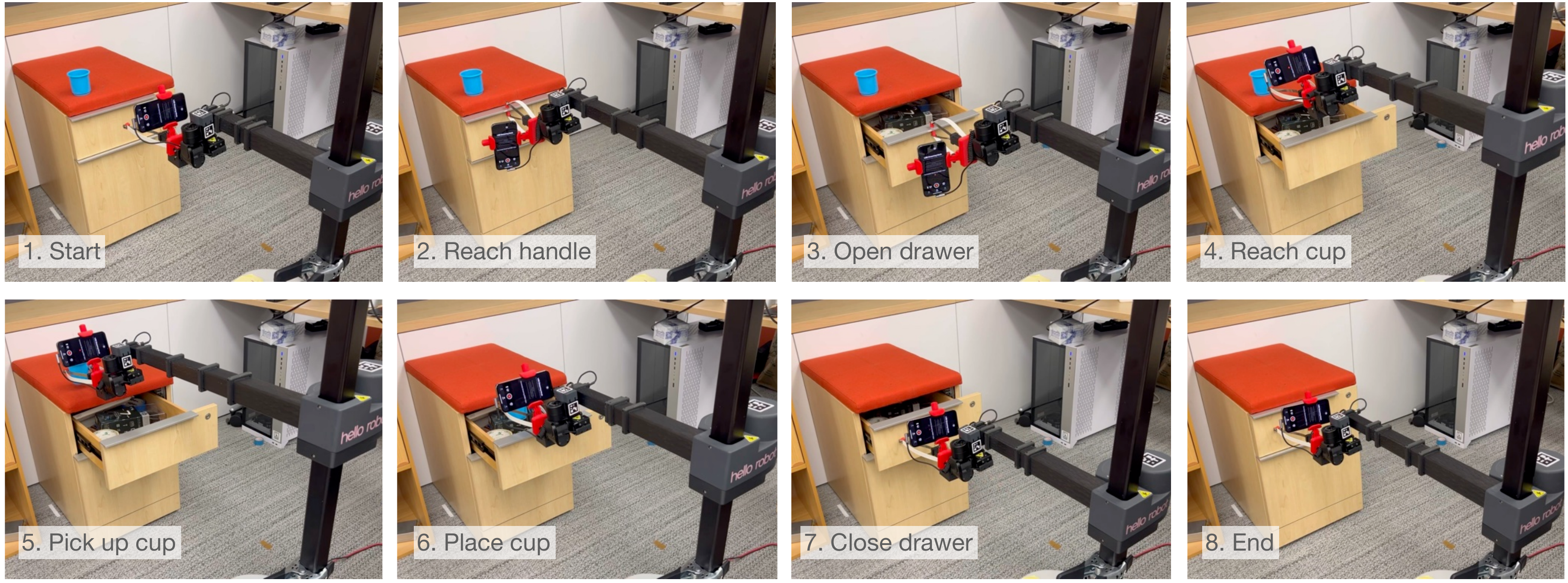}
        \caption{The robot opening a drawer, placing a cup inside of it, and closing it afterwards.}
        \label{fig:long-horizon-cup-drawer}
    \end{subfigure}
    
    \vspace{1em} %
    
    \begin{subfigure}{\textwidth}
        \centering
        \includegraphics[width=\linewidth]{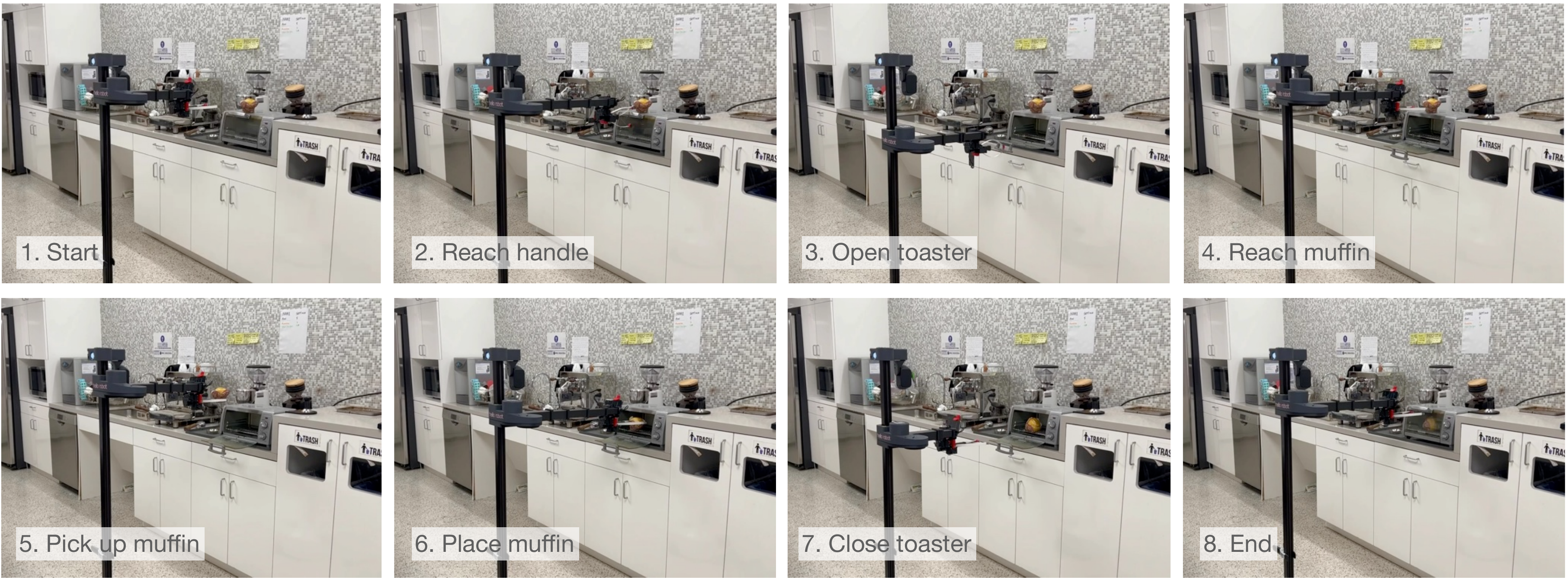}
        \caption{The robot opening a toaster oven, placing a muffin inside of it, and closing it.}\label{fig:long-horizon-muffin-toaster}
    \end{subfigure}
    
    \vspace{1em} %
    
    \begin{subfigure}{\textwidth}
        \centering
        \includegraphics[width=\linewidth]{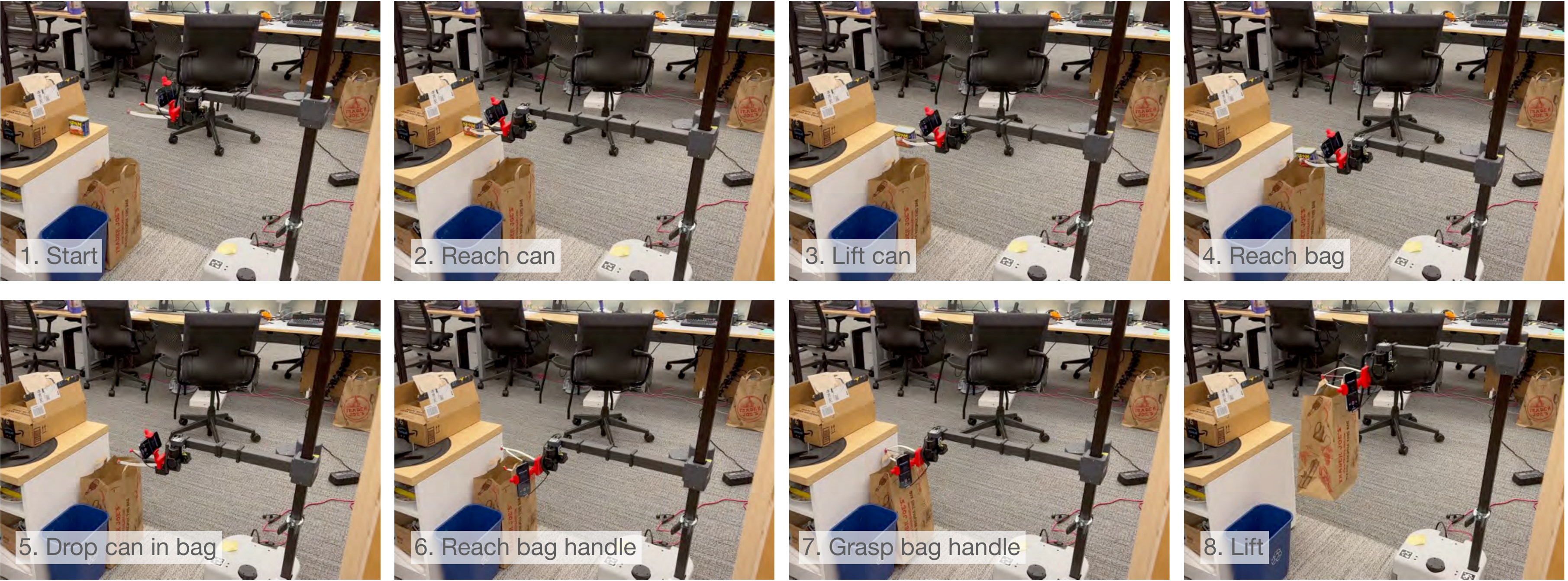}
        \caption{The robot picking up a can, placing it in a bag, and then lifting it.}
        \label{fig:long-horizon-can-bag}
    \end{subfigure}
    \caption{\method{} completing three temporally extended tasks each made up of five to seven subtasks.}
\end{figure}

\subsection{Scaling to Long Horizon Tasks}
\label{sec:long-horizon}
We primarily focused on short-horizon tasks in this work, but intuitively, our framework should be easily extensible to longer-horizon, multi-step tasks with algorithmic improvements.
To validate this intuition, we train \method{} to perform some multi-step tasks in our lab.

In Figures~\ref{fig:long-horizon-cup-drawer},~\ref{fig:long-horizon-muffin-toaster}, and ~\ref{fig:long-horizon-can-bag}, we can see that~\method{} can successfully perform multi-step, long horizon tasks like putting a cup in a drawer, placing a muffin in a toaster oven, or placing a can in a recycling bag and lifting it.
However, because of the compound nature of these tasks, the failure cases also tend to compound with our simple methods, as seen in Figure~\ref{fig:long-horizon-analysis}.
For example, in the muffin-in-toaster task, our model got 1 success out of 10 trials, and in the cup-in-drawer task, our model got 6 success out of 10 trials.
In both cases, the sub-task causing primary failure was not letting go of the grasped object (cup or muffin).
If we can improve on such particular subtasks, possibly using force-aware methods similar to~\cite{collins2023forcesight}, we believe \method{} can easily scale up to long-horizon tasks.
Fast on-line adaptation on top of offline training~\cite{haldar2023rot, haldar2023fish} has potential to improve such long horizon cases as well.
In other cases, the robot was able to open the door but unable to disengage safely from the handle because some part of the robot gripper got stuck to the handle.
This failure mode points to the need of better designed, less bare-boned robot grippers for household tasks.

\subsection{Incorporating Memory}
Another large challenge in our setup is the problem of robotic scene memory.
With a single first person point of view on the \stick{}, the robot needs to either see or remember large parts of the scene to operate on it effectively. However, there is a dearth of algorithms that can act as standalone memory module for robots. The algorithms that currently exist, such as~\cite{shafiullah2022clip, kerr2023lerf, rashid2023lerftogo, wang2023d3, shen2023distilled, jatavallabhula2023conceptfusion, bolte2023usa, huang2023visual} also tend to have a rigid representation of the scene that is hard to change or edit on the fly, which will need to improve for real household deployments.

\subsection{Improving Sensors and Sensory Representations}
Most of current visual representation learning algorithms focus on learning from third-person views, since that is the dominant framework in Computer Vision. However, third person cameras often rely on  camera calibration, which generally makes using large robot datasets and transferring data between robots difficult~\cite{bharadhwaj2023roboagent}. A closer focus on learning from first person cameras and eye-in-hand cameras would make sharing data from different environments, tasks, and robots much easier. 
Finally, one of the modality that our \stick{} is missing is having tactile and force sensors on the gripper. In deployment, we have observed the robot sometimes applies too much or too little force because our framework doesn't contain such sensors. Better integration of cheap sensors~\cite{bhirangi2021reskin} with simple data collection tools like the \stick{}, or even more methods like learned visual contact force estimation~\cite{grady2022visual, collins2023visual} could be crucial in such settings.

\subsection{Robustifying Robot Hardware}
A large limitation on any home robotics project is the availability of cheap and versatile robot platforms. While we are able to teach the Hello Robot Stretch a wide-variety of tasks, there were many more tasks that we could not attempt given the physical limitations of the robot: its height, maximum force output, or dexterous capabilities. Some of these tasks may be possible while teleoperating the robot directly rather than using the \stick, since the demonstrator can be creative and work around the limits. However, availability of various home-ready robotic platforms and further development of such demonstration tools would go a long way to accelerate the creation of household robot algorithms and frameworks.

\section{Reproducibility and Call for Collaboration}
\label{sec:reproducibility}
\newcommand{\hardwaregithub}{https://github.com/notmahi/dobb-e/tree/main/hardware}
\newcommand{\modeltraininggithub}{https://github.com/notmahi/dobb-e/tree/main/imitation-in-homes}
\newcommand{\dataprocessinggithub}{https://github.com/notmahi/dobb-e/tree/main/stick-data-collection}
\newcommand{\robotcodegithub}{https://github.com/notmahi/dobb-e/tree/main/robot-server}
\newcommand{\documentation}{https://docs.dobb-e.com}
\newcommand{\huggingface}{https://huggingface.co/notmahi/dobb-e}

To make progress in home robotics it is essential for research projects to contribute back to the pool of shared knowledge. To this end, we have open-sourced practically every piece of this project, including hardware designs, code, dataset, and models. Our primary source of documentation for getting started with \method{} can be found at \url{\documentation}.

\begin{itemize}[leftmargin=12pt]
    \item \textbf{Robot base:} Our project uses Hello Robot Stretch as a platform, which is similarly open sourced and commercially available on the market for US\$24,000 as of November 2023.
    \item \textbf{Hardware design:} We have shared our 3D-printable STL files for the gripper and robot attachment in the GitHub repo: \url{\hardwaregithub}. We have also created some tutorial videos on putting the pieces together and shared them on our website. The reacher-grabber stick can be bought at online retailers, links to which are also shared on our website \url{\website /#hardware}.
    \item \textbf{Dataset:} Our collected home dataset is shared on our website. We share two versions, a 814 MB version with the RGB videos and the actions, and an 77 GB version with RGB, depth, and the actions. They can be downloaded from our website, \url{\website /#dataset}. At the same time, we share our dataset preprocessing code in GitHub \url{\dataprocessinggithub} so that anyone can export their collected R3D files to the same format.
    \item \textbf{Pretrained model:} We have shared our visual pretraining code as well as checkpoints of our pretrained visual model in our GitHub \url{\modeltraininggithub} and Huggingface Hub \url{\huggingface}. For this work, we also created a high efficiency video dataloader for robotic workload, which is also shared under the same GitHub repository.
    \item \textbf{Robot deployment:} We have shared our pretrained model fine-tuning code in \url{\modeltraininggithub}, and the robot controller code in \url{\robotcodegithub}. We also shared a step-by-step guide to deploying this system in a household, as well as best practices that we found during our experiments, in a handbook under \url{\documentation}.
\end{itemize}

Beyond these shared resources, we are also happy to help other researchers set up this framework in their own labs or homes. We have set up a form on our website to schedule 30-minutes online meetings, and shared some available calendar slots where we would be available to meet online and help set up this system. We hoping these steps would be beneficial for practitioners to quickly get started with our framework.

Finally, we believe that our work is an early step towards learned household robots, and thus can be improved in many possible ways. So, we welcome contributions to our repositories and our datasets, and invite researchers to contact us with their contributions. We would be happy to share such contributions with the world with proper credits given to the contributors.

\begin{ack}

NYU authors are supported by grants from Amazon, Honda, and ONR award numbers N00014-21-1-2404 and N00014-21-1-2758. NMS is supported by the Apple Scholar in AI/ML Fellowship. LP is supported by the Packard Fellowship. Our utmost gratitude goes to our friends and colleagues who helped us by hosting our experiments in their homes, and those who helped us collect the pretraining data. We thank Binit Shah and Blaine Matulevich for support on the Hello Robot Platform and the NYU HPC team, especially Shenglong Wang, for compute support. We thank Jyo Pari and Anya Zorin for their work on earlier iterations of the~\stick{}. We additionally thank Sandeep Menon and Steve Hai for his help in the early stages of data collection. We thank Paula Nina and Alexa Gross for their input on the designs and visuals. We thank Chris Paxton, Ken Goldberg, Aaron Edsinger, and Charlie Kemp for feedback on early versions of this work. Finally, we thank Zichen Jeff Cui, Siddhant Haldar, Ulyana Pieterberg, Ben Evans, and Darcy Tang for the valuable conversations that pushed this work forward.
\end{ack}

\newpage
\bibliographystyle{unsrt}
\bibliography{references}

\clearpage
\appendix

\newpage

\clearpage

\end{document}